\documentclass[conference]{IEEEtran}
\IEEEoverridecommandlockouts
\usepackage{cite}
\usepackage{amsmath,amssymb,amsfonts}
\usepackage{algorithmic}
\usepackage{graphicx}
\usepackage{textcomp}
\usepackage{xcolor}
\usepackage{url}
\usepackage[export]{adjustbox}
\usepackage{newtxmath}
\ifCLASSOPTIONcompsoc
\usepackage[caption=false,font=normalsize,labelfont=sf,textfont=sf]{subfig}
\else
\usepackage[caption=false,font=footnotesize]{subfig}
\fi
\usepackage{todonotes}
\def\BibTeX{{\rm B\kern-.05em{\sc i\kern-.025em b}\kern-.08em
    T\kern-.1667em\lower.7ex\hbox{E}\kern-.125emX}}
\begin{document}

\title{Learning Perceptive Bipedal Locomotion over Irregular Terrain\\
}

\author{\IEEEauthorblockN{Bart van Marum, Matthia Sabatelli, Hamidreza Kasaei}
\IEEEauthorblockA{\textit{Department of Artificial Intelligence} \\
\textit{University of Groningen}\\
bart@vmeps.com, m.sabatelli@rug.nl,  hamidreza.kasaei@rug.nl 
}
}

\maketitle
\begin{abstract}
In this paper we propose a novel bipedal locomotion controller that uses noisy exteroception to traverse a wide variety of terrains.
Building on the cutting-edge advancements in attention based belief encoding for quadrupedal locomotion, our work extends these methods to the bipedal domain, resulting in a robust and reliable internal belief of the terrain ahead despite noisy sensor inputs.
Additionally, we present a reward function that allows the controller to successfully traverse irregular terrain. 
We compare our method with a proprioceptive baseline and show that our method is able to traverse a wide variety of terrains and greatly outperforms the state-of-the-art in terms of robustness, speed and efficiency.
\end{abstract}


\section{Introduction}
Humanoid robots hold immense potential as a general-purpose platform for various applications due to their compatibility with human-designed environments. This compatibility enables humanoid robots to seamlessly work alongside humans, reducing the need for expensive modifications to existing infrastructure. Despite the benefits, creating a fully functional and general-purpose humanoid robot still poses several technical challenges, including locomotion over irregular and previously unseen terrain. To address this challenge, the present work focuses on developing a robust and reliable bipedal locomotion controller.

Conventionally bipedal locomotion controllers are designed as complicated state machines and explicit dynamical models~\cite{8796090,cassie-michigan}. However, these models lack in robustness, do not generalize to new scenarios or terrains without explicit modelling, and are laborious and complicated to develop and maintain. Moreover, adding exteroceptive capabilities to such methods is not straightforward.

In recent years, there has been a shift towards the use of Reinforcement Learning (RL)~\cite{Sutton1998} based controllers for simulated~\cite{Peng_2017, 10.1145/3197517.3201397}, as well as real-world bipedal robots~\cite{Xie_Berseth,berkely_cassie,siekmann_godse,DBLP:journals/corr/abs-2006-02402,Siekmann-RSS-21}. These methods model the control policy as a neural network and train them to maximize some reward signal. This approach has proven to be robust, even in the face of motor malfunctions~\cite{berkely_cassie}.

Many RL based approaches rely on the use of reference trajectories and imitation rewards to train a policy to produce a gait\cite{Xie_Berseth}, limiting the policy to learn predetermined behaviour. However, recent work suggests that it is possible to learn a wide variety of different bipedal gaits in a single neural network by using periodic reward functions~\cite{siekmann_godse}.
\begin{figure}[t]
	\centering
	\includegraphics[width=\linewidth]{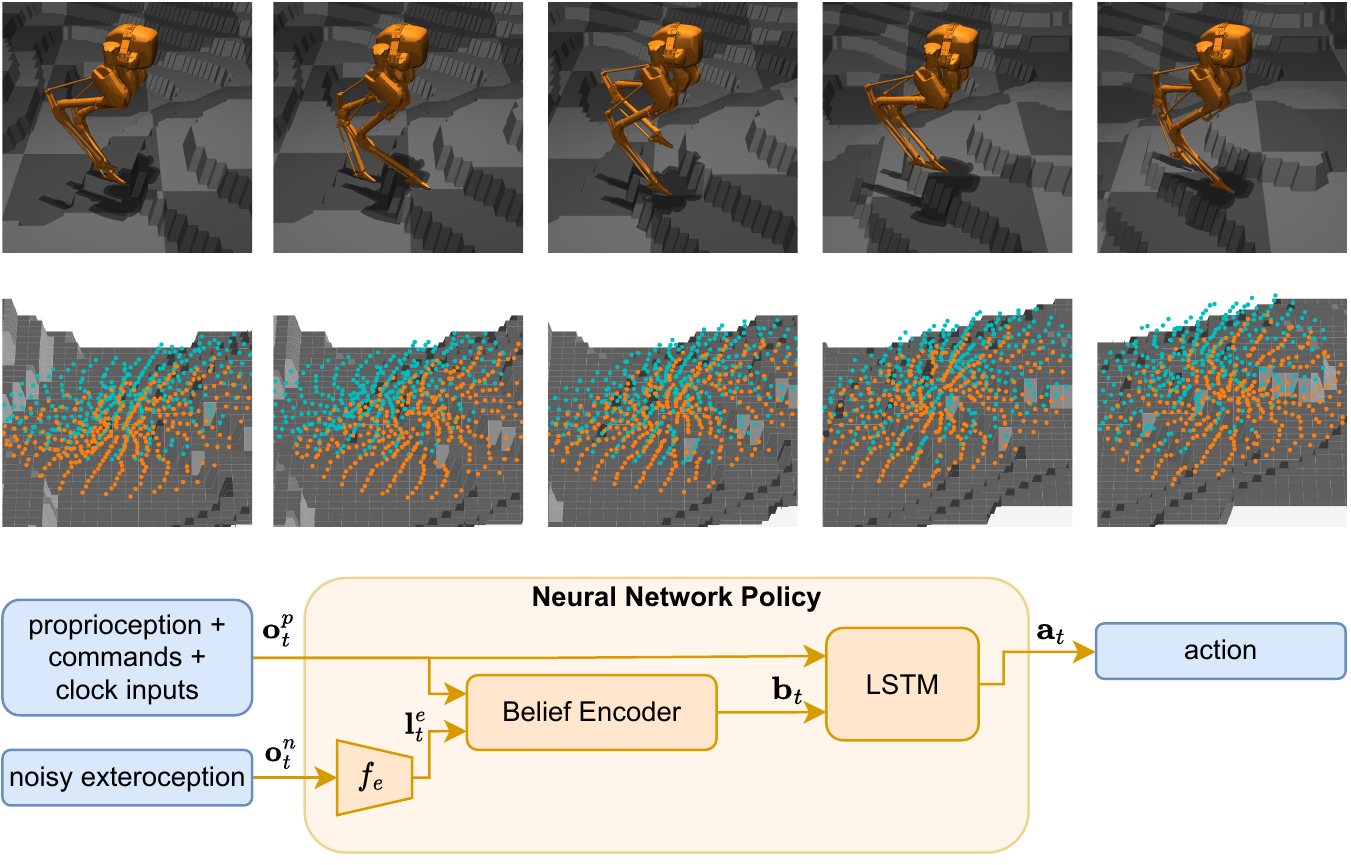}
	\caption{
		In this work we develop a bipedal locomotion control policy based on both exteroception and proprioception that is able to traverse a wide variety of terrains.
		The first row shows Cassie walking over random terrain in the physics simulator.
		The second row shows noisy exteroceptive samples that are input to the policy at the same timesteps.
		The bottom shows the policy architecture during inference.
	}
	\label{fig:fig1}
\end{figure}
Despite the great progress in recent years for neural network based bipedal locomotion controllers, most approaches are compatible with flat terrain only. Some successful attempts have been made to learn blind bipedal locomotion controllers for more challenging terrains~\cite{Siekmann-RSS-21}, however, these policies resort to more conservative foot trajectories with higher steps and are unable to avoid dangerous areas. Such blind strategies do not generalize well to a wide variety of unseen and irregular terrains and lack in efficiency, and therefore are not feasible for a fully capable humanoid robot.

In order for a legged locomotion controller to traverse any random previously unseen terrain robustly, it needs information about the world ahead. A controller that is based on proprioception only is limited to reactive and precautionary behaviour. Only a controller that has information about the world ahead can actively plan steps. Exteroceptive inputs are necessary, however using both proprioception and exteroception presents a challenge. Exteroceptive sensors such as cameras, Lidar, or radar often produce spurious readings in cases such as reflection (water puddle), transparency (glass), deformation (snow), or fake obstacles (tall grass). This may lead locomotion policies based on such inputs to unnecessarily avoid certain areas or fail outright, and raises the question of how to handle disagreeing proprioceptive data.

The field of quadrupedal locomotion control has shown great progress in recent years in learning controllers for navigating challenging terrain~\cite{Lee_2020, rudin2021parallel}. Most notably,~\cite{Miki_2022} shows that using a recurrent belief encoder with an attention mechanism, a neural network policy is able to learn when to trust and when not to trust the exteroceptive data. This allows the locomotion controller to utilize exteroceptive data when it is most useful, and fall back to proprioceptive data when it is not. Our work extends these methods to the bipedal domain, resulting in a robust and reliable internal belief of the terrain ahead despite noisy sensor inputs.

\subsubsection*{Contribution}
In this work we apply a recurrent attention based belief encoder to a bipedal locomotion policy to develop a robust controller capable of traversing irregular terrain based on noisy exteroception. We present a reward function leveraging prior work that allows the policy to learn traversing rough terrain. 
We perform a wide range of simulation based experiments to show that our controller is able to navigate a variety of terrains while outperforming state-of-the-art proprioceptive controllers in terms of robustness, efficiency, and speed.
Figure~\ref{fig:fig1} shows the architecture of our controller. 

The remainder of the paper is organized as follows: The next section describes the methods used to train our controller. Section~\ref{sec:experimental_results} describes the training process and the experiments performed to evaluate the performance of our controller, including the results. Finally, Section~\ref{sec:discussion} concludes the paper.

\section{Methods}
\label{sec:methods}
\subsection{Learning Setup}
The main goal is to develop a robust bipedal locomotion controller that is able to navigate irregular terrain while following a command. In order to do so we use privileged learning~\cite{privileged_learning} to distill a policy that is able to work with potentially noisy and spurious exteroceptive observations. Previous work has shown that directly learning the desired behavior over rough terrain with RL does not converge within reasonable time budgets~\cite{Lee_2020}. First a teacher policy with access to perfect, noise free observations is trained in simulation through reinforcement learning to traverse a wide range of different terrains. We then train a student policy to imitate the behaviour of the teacher policy, but without privileged information and noisy inputs. We use Proximal Policy Optimization (PPO)~\cite{ppo} to train our policies, as PPO has shown to yield good results in bipedal locomotion control~\cite{berkely_cassie, siekmann_godse, Xie_Berseth}.

\subsection{State and Action Representation}
We define three observations $\mathbf{o}^p_t, o^e_t, o^n_t$. Here $\mathbf{o}^p_t \in \mathbb{R}^{44}$ is the proprioceptive input, consisting of motor positions, motor speeds, joint positions, joints speeds, pelvis orientation, pelvis angular velocity, user commanded velocity $v_c$ and clock inputs $i_t$. The user commanded velocity $v_c$ is defined as the pair $(\mathbf{v}_{cmd}, \omega_{cmd})$  where $\mathbf{v}_{cmd} \in \mathbb{R}^2$ represents the commanded velocity in the $x$ and $y$ directions and $\omega_{cmd}$ represents the commanded angular velocity around the $z$ axis. The clock input $i_t$ is defined as the pair $\sin(2\pi(\phi))$ and $\sin(2\pi(\phi + 0.5))$, where $\phi$ is defined as $\phi = (t \mod T)/T$, with $t$ denoting the current timestep, and $T$ a user defined gait period in terms of timesteps.
Although it has been noted in past research~\cite{DBLP:journals/corr/abs-2006-02402} that clock inputs are necessary, we found in preliminary experiments that a policy without clock inputs learns a gait with no meaningful effect on the reward. However, we have not yet performed a thorough investigation on this matter.

\begin{figure}[tbp]
	\centering
	\begin{tabular}{cc}
		\includegraphics[trim=1430 670 1350 670, clip, width=0.48\linewidth, valign=t]{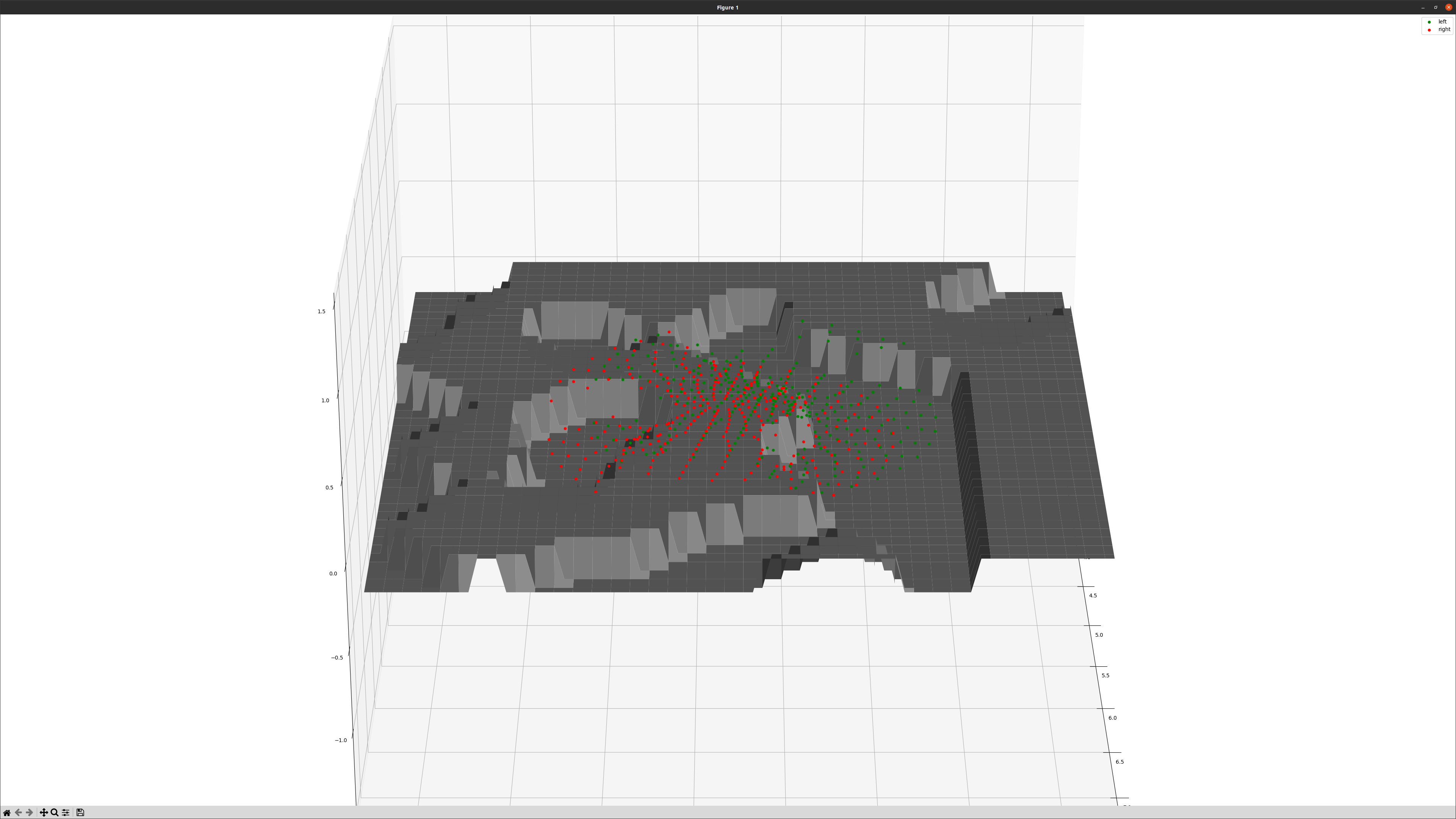} &
		\includegraphics[trim=0 0 0 0, clip, width=0.48\linewidth, valign=t]{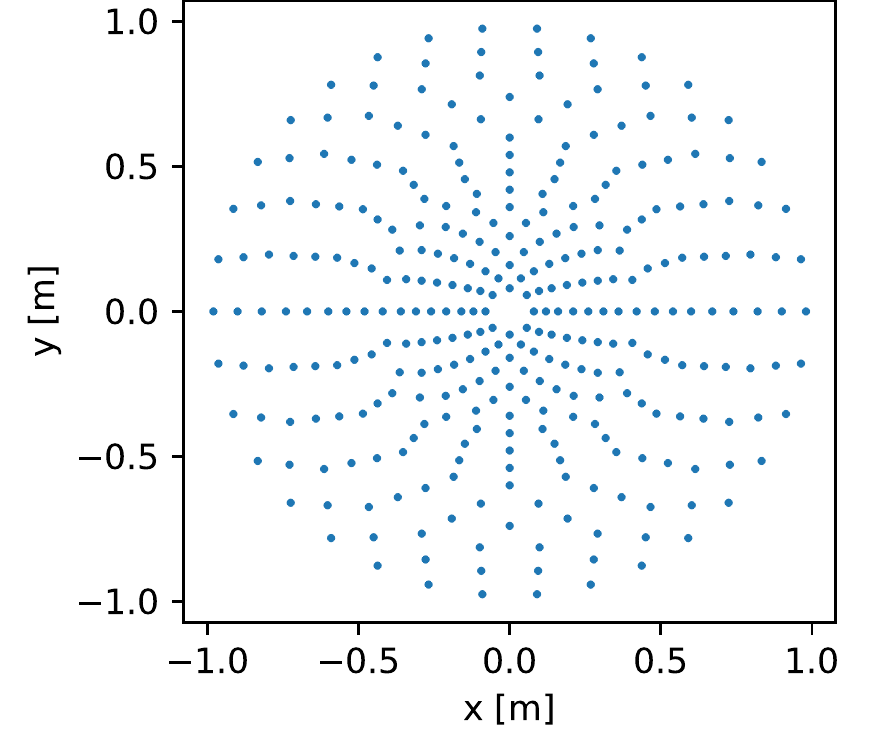} \\
		(a) & (b) \\
	\end{tabular}
	\caption{(a) A close up view of the exteroceptive simulator. The red dots represent the height sample for the right foot, and the green dots for the left foot. Samples are taken with the sampling pattern centered around the $xy$ position of each foot and rotated to match pelvis orientation. (b) Shows a detailed plot of the pattern used to sample heights from the terrain.}
	\label{fig:exteroceptor}
\end{figure}

Additionally, $o^e_t = (\mathbf{e}^l_t, \mathbf{e}^r_t)$ is the pair of noiseless exteroceptive observations for the left and right feet. To obtain the observations the terrain height is sampled with a sampling pattern centered at the location of the respective foot. The sampling pattern consists of 318 points spaced circularly. Figure~\ref{fig:exteroceptor} shows the height sampling taking place in the simulation environment. 

Finally, $o^n_t = (\mathbf{n}^l_t, \mathbf{n}^r_t)$ is the pair of noisy exteroceptive observations. Sampling is similar to $\mathbf{o}^e_t $ however a noise is applied to the sampling pattern coordinates and sampled values. Further details on noise generation are discussed in section~\ref{subsec:randomization}. 

The action $\mathbf{a}_t \in \mathbb{R}^{10}$ represents the PD targets for the actuators in the robot model. Previous research has shown that PD targets are an effective action parametrization for learning locomotion~\cite{peng_pd_targets}. The robot PD controller runs at 2 kHz, whereas actions are sampled from the policy at a rate of 40 Hz.

\subsection{Policy Architecture}
\begin{figure}[tbp]
	\centerline{\includegraphics[trim=77 230 129 435, clip, width=\linewidth]{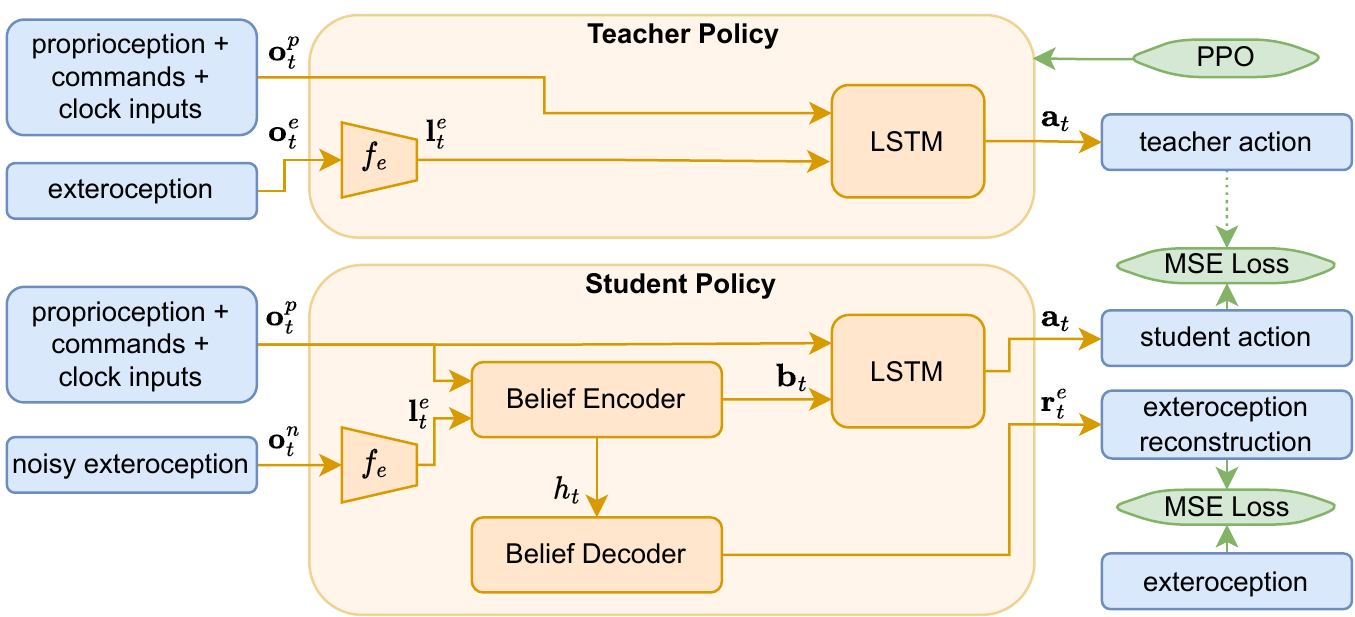}}
	\caption{The top figure shows the teacher policy architecture which is trained with PPO. The bottom figure shows the student policy architecture which is trained to both imitate the action output of the teacher policy, and to denoise the noisy exteroceptive input.}
	\label{fig:policy_architecture}
\end{figure}

The teacher and student policy architectures are illustrated in Figure~\ref{fig:policy_architecture}. In this section, we will provide a detailed description of these architectures.

\begin{figure*}[t]
	\centering
	\subfloat[]{
		\includegraphics[trim=1100 300 1100 300, clip, width=0.19\linewidth]{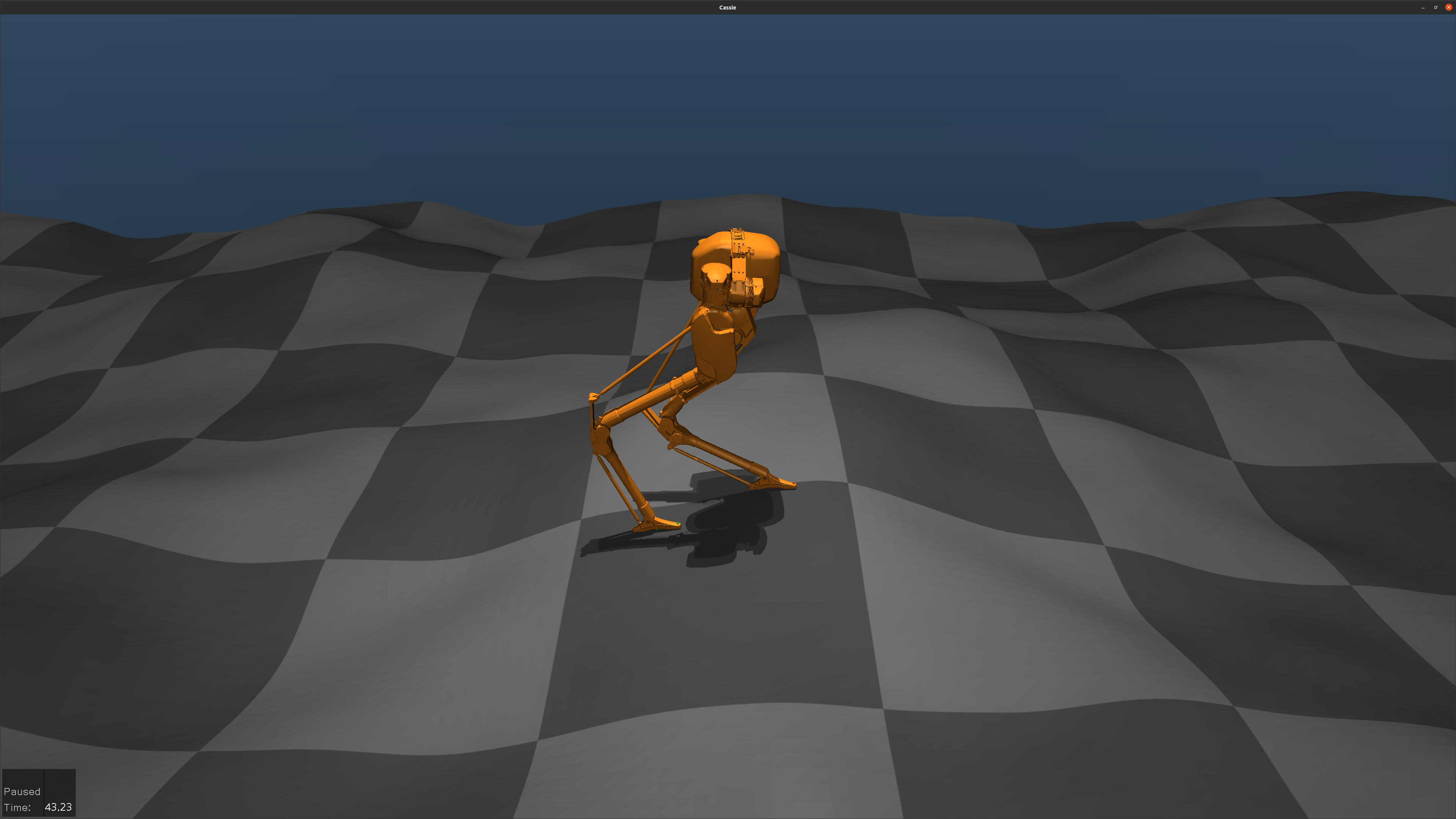}
	}
	\subfloat[]{
		\includegraphics[trim=1100 300 1100 300, clip, width=0.19\linewidth]{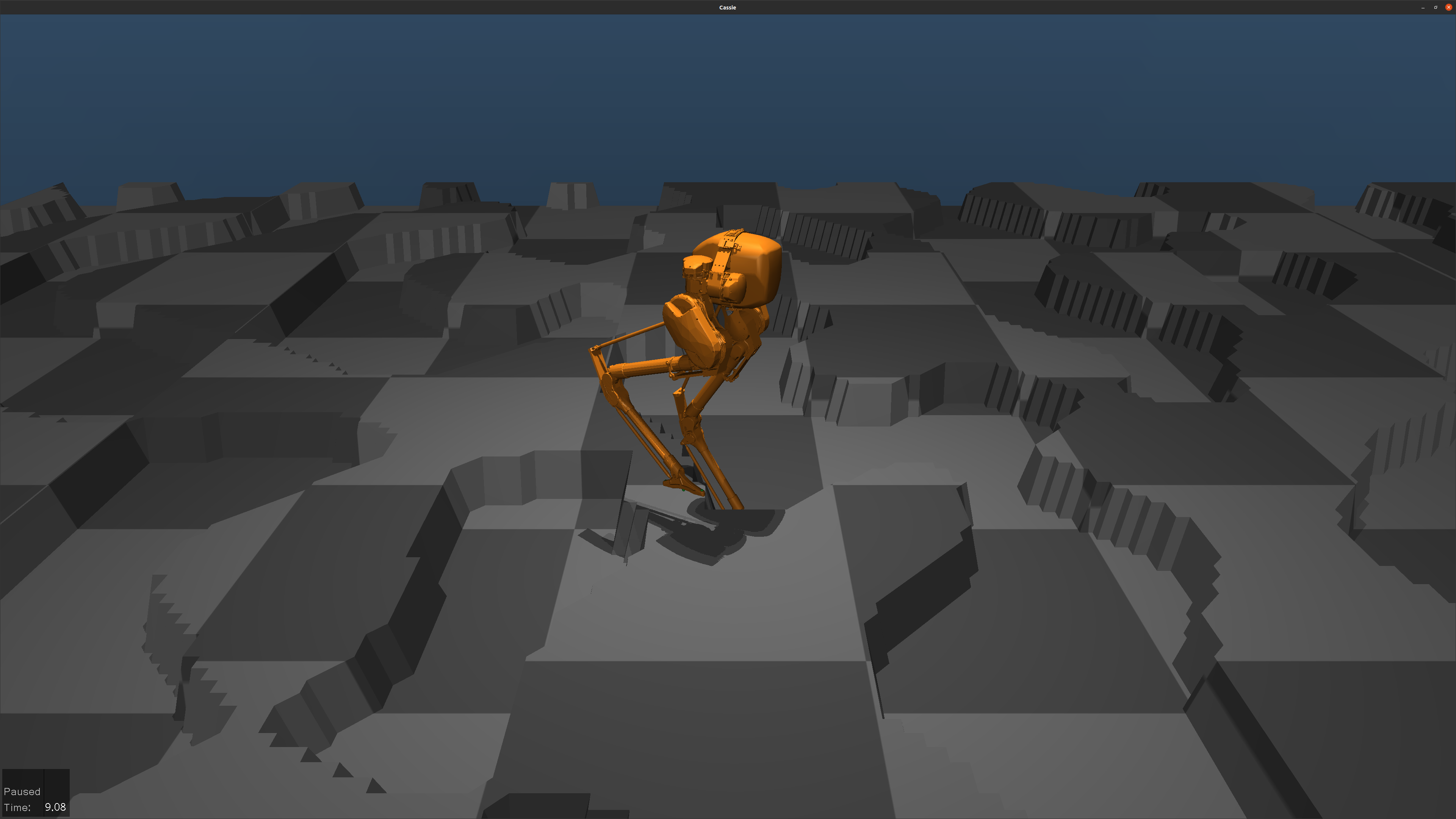}
	}
	\subfloat[]{
		\includegraphics[trim=1100 300 1100 300, clip, width=0.19\linewidth]{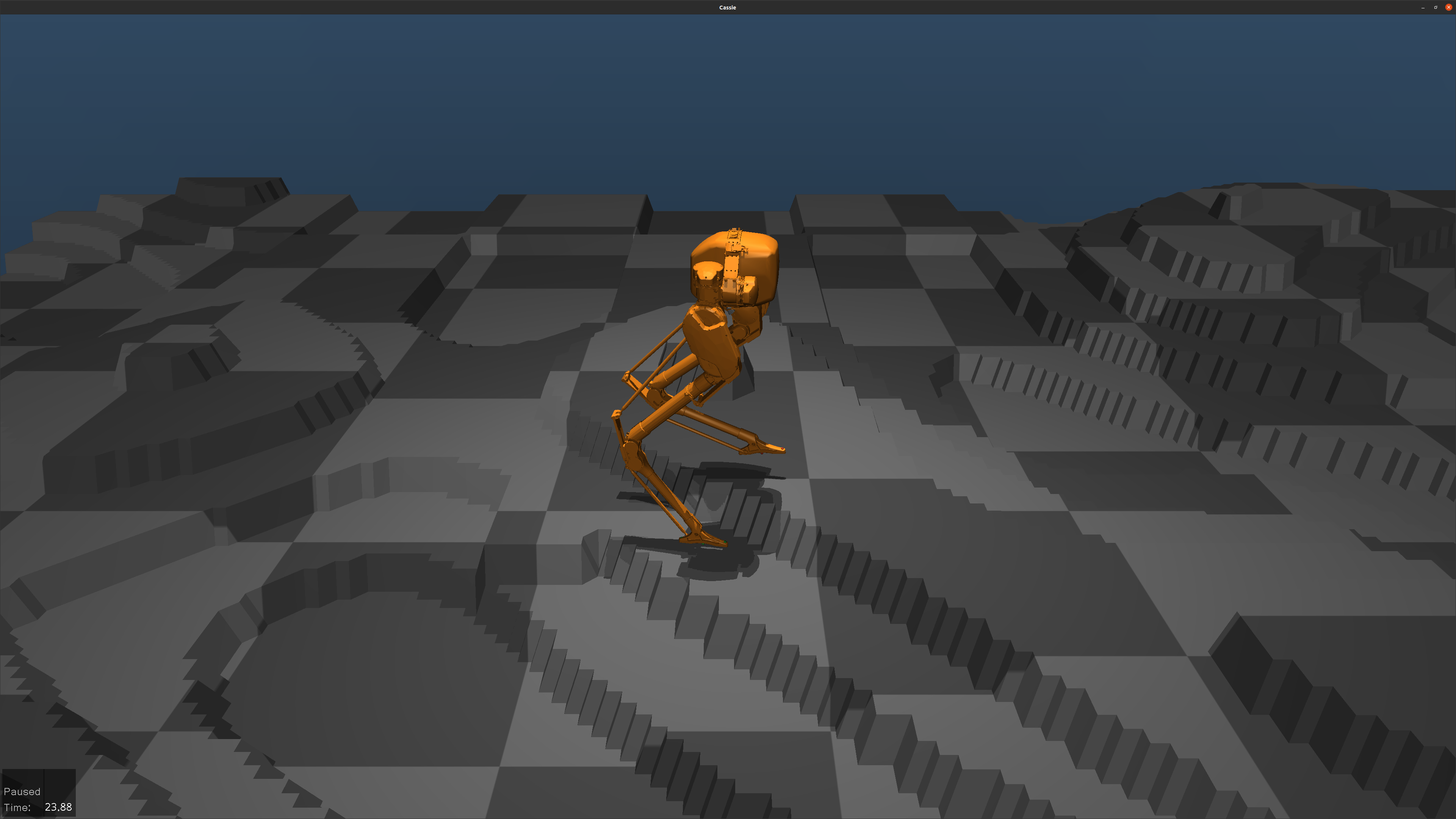}
	}
	\subfloat[]{
		\includegraphics[trim=1100 300 1100 300, clip, width=0.19\linewidth]{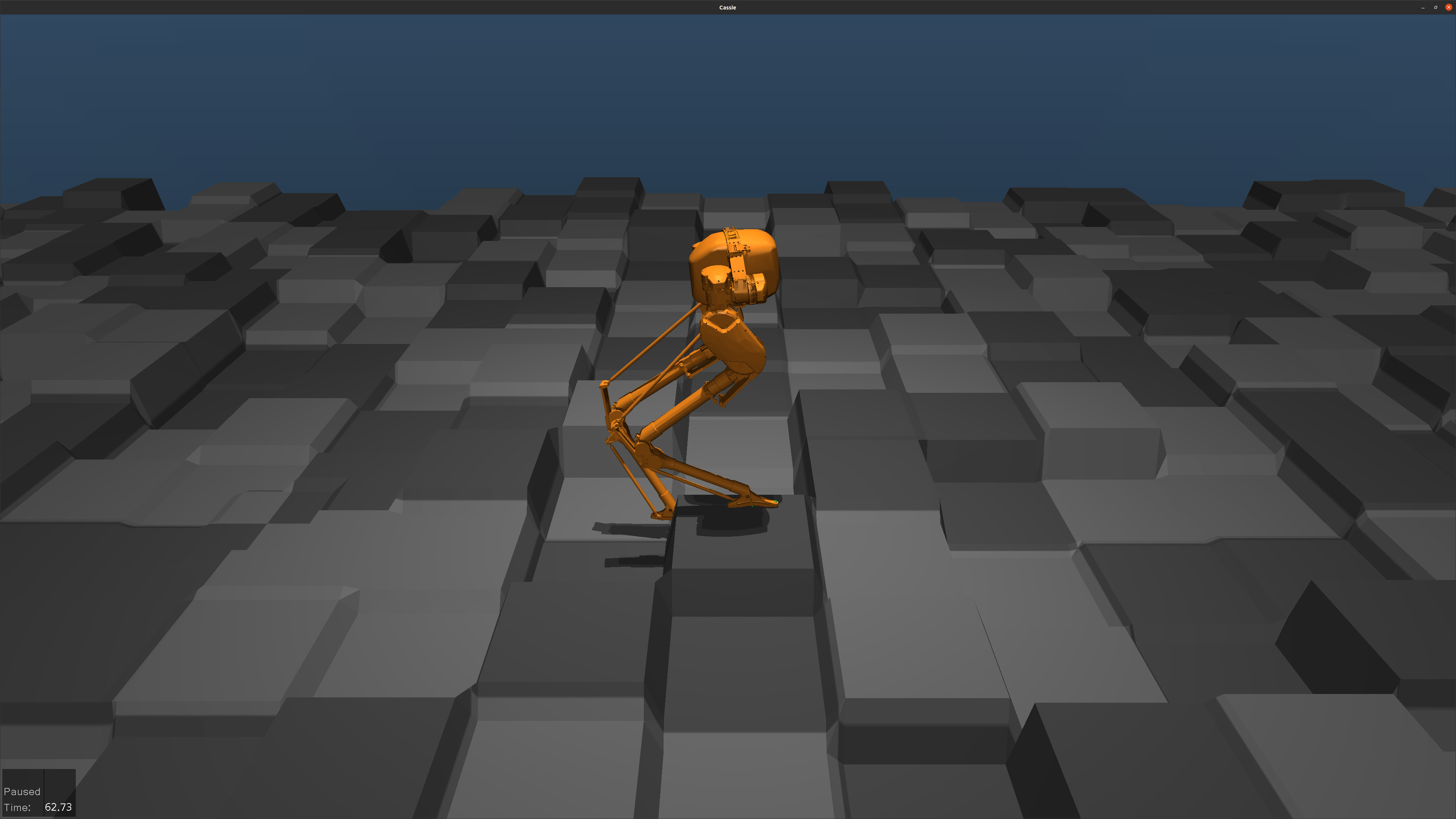}
	}
	\subfloat[]{
		\includegraphics[trim=1100 300 1100 300, clip, width=0.19\linewidth]{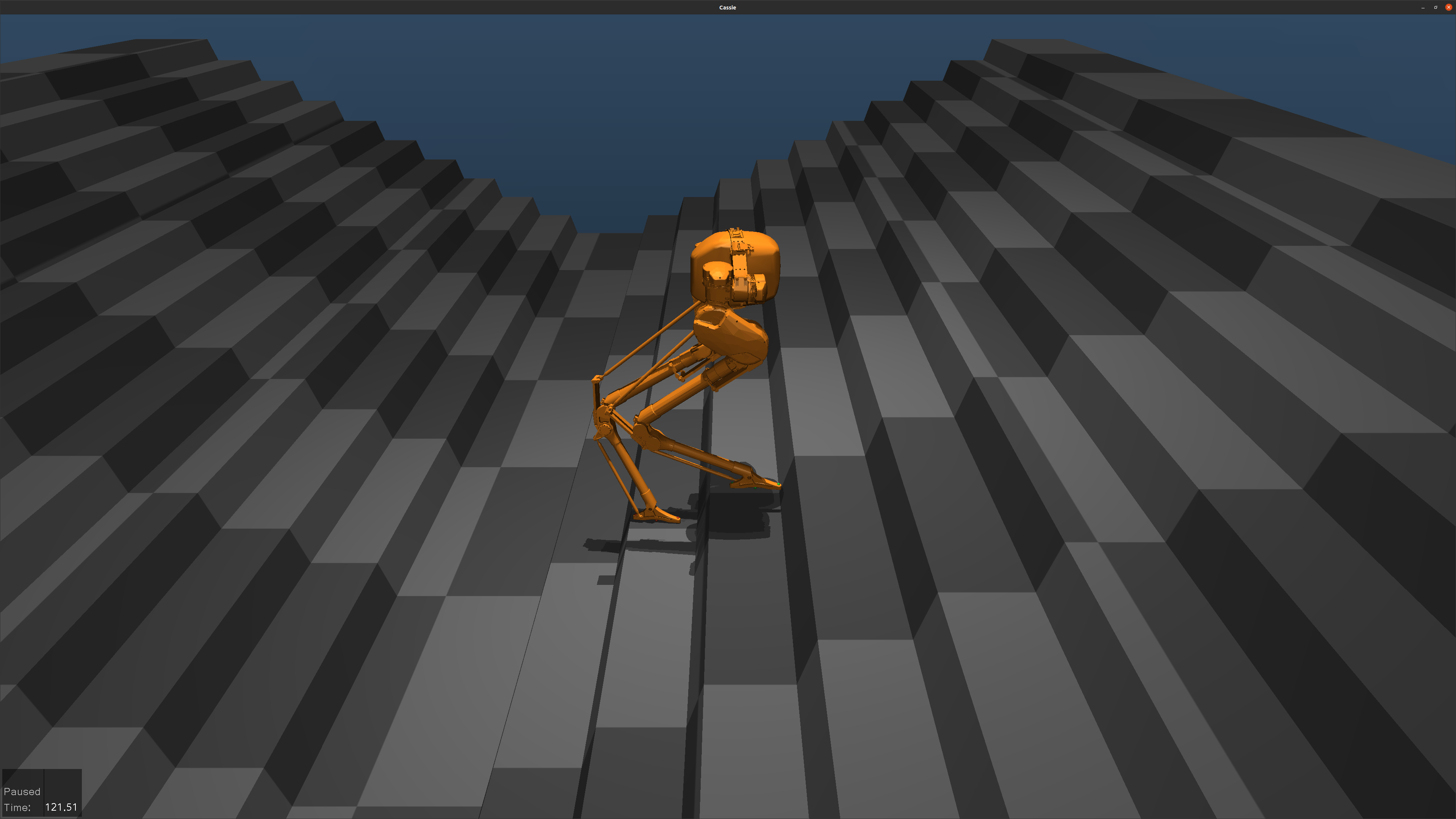}
	}
	\caption{The five different terrain modes used for training: (a) \textit{hills}, (b) \textit{edges}, (c) \textit{quantized hills}, (d) \textit{squares}, (e) \textit{stairs}.}
	\label{fig:terrain}
\end{figure*}

\subsubsection{Teacher Policy}
The teacher policy $\pi^t$ consists of an exteroceptive encoder $f_e$ and an LSTM~\cite{hochreiter1997long}. The encoder $f_e$ consists of 3 fully connected layers of size \{256, 160, 96\} and the LSTM has two layers of 256 nodes. The teacher policy receives the observation $o^t_t = (\mathbf{o}_t^p, \mathbf{o}_t^e)$. The exteroceptive encoder $f_e$ receives both exteroceptive observations $\mathbf{e}_t^{l}, \mathbf{e}_t^{r}$ that are in $o_t^e$ and encodes them separately into the latent vectors $\mathbf{l}_t^{el}$ and $\mathbf{l}_t^{er}$ which are concatenated into the latent vector $\mathbf{l}_t^e \in \mathbb{R}^{200}$. The LSTM receives the concatenation of $\mathbf{o}_t^p$ and $\mathbf{l}_t^e$ and outputs an action $\mathbf{a}_t$.

\subsubsection{Student Policy}
The student policy $\pi^s$ takes in the noisy observation $o^s_t = (\mathbf{o}_t^p, \mathbf{o}_t^n)$ and has a partially similar architecture as the teacher policy, using the same encoder $f_e$ and LSTM. An added component is a recurrent belief encoder, which receives the exteroceptive latent vector $\mathbf{l}^e_t$ and the proprioceptive observation $\mathbf{o}_t^p$ and outputs a belief vector $\mathbf{b}_t \in \mathbb{R}^{192}$. The belief vector is then concatenated with the proprioceptive observation $\mathbf{o}_t^p$ and fed into the LSTM, which in turn outputs an action $\mathbf{a}_t$.

The main aspect of the student policy is the recurrent belief encoder, an approach introduced by~\cite{Miki_2022}, which is intended to take the proprioception and noisy exteroception and develop an internal representation of what the terrain looks like. In order to train this internal representation a belief decoder is added to the policy, which takes as input the hidden state of the recurrent belief encoder. The belief decoder outputs a reconstruction of the exteroceptive inputs, which is trained to minimize the difference with the noise free exteroceptive observation $o^e_t$. This method encourages the internal hidden state of the belief encoder to represent a representation of the outside world that is as accurate as possible, despite noisy inputs. Additionally, the belief encoding system is fitted with an attention mechanism, such that the policy is able to learn when exteroceptive data is not useful, and rely on proprioception instead to construct the belief. For more detail about the belief encoder and decoder we refer the reader to~\cite{Miki_2022}.

The student policy is trained with both an action imitation loss, and an observation reconstruction loss to encourage the internal belief representation of the outside world to be as accurate as possible. 

\subsection{Terrain Generation}

We use a linear curriculum to ramp terrain generation intensity. The ramp starts after the policy has learned to walk on flat ground. All generated terrains are modelled as a height map in meters and multiplied with the curriculum factor $c_t \in [0,1]$. 

We define five different terrain modes, as shown in Figure~\ref{fig:terrain}. 
The first is \textit{hills}, which is modelled as a sum of a low frequency and a higher frequency Perlin noise~\cite{perlin-noise}. The generated values are normalized to a range $[0, 0.8]$. 
The second terrain mode is \textit{edges}, which consists of a Perlin noise that has been quantized to two levels~${\{0, h \sim \mathcal{U}(0.15, 0.25)\}}$. 
The third mode is  \textit{squares} which consists of a grid of squares with sides $d \in [0.4, 0.6]$ of random height~$h \in [0, 0.4]$. 
The fourth mode is \textit{quantized hills} which is Perlin noise that has been quantized to discrete levels with a random step size~$h \in [0.12, 0.18]$. 
The fifth and final mode of terrain generation is \textit{stairs}, consisting of alternating ascending and descending staircases. To generate a staircase we randomly select a run~$d \in [0.3, 0.4]$ and rise~$r \in [0.1, 0.22]$ for 10 equal stairs. 

\subsection{Randomization}
\label{subsec:randomization}
Although we only focus on simulation based experiments in this work, previous work has shown that policies trained in simulation are able to bridge the sim-to-real gap given proper domain randomization~\cite{DBLP:journals/corr/abs-2006-02402, domainRandomization}. Therefore we randomize joint damping, body part masses and friction coefficients at the start of each episode. We use the same parametrization as presented in~\cite{Siekmann-RSS-21}.

Furthermore, we randomize the velocity command to expose the policy to a range of different velocity commands during training. At the start of each episode and at one random timestep during each episode a new velocity command is sampled. The probability distribution for the velocity commands is shown in Table~\ref{tab:vel_commands}.

\begin{table}[ht]
	\renewcommand{\arraystretch}{1.3}
	\caption{Velocity Commands Randomization}
	\label{tab:vel_commands}
	\centering
	\begin{tabular}{l|l|l||l}
	\hline
	$v_x$ & $v_y$ & $\omega_z$ & \bfseries Probability\\
	\hline\hline
	$0$ & $0$ & $0$ & $0.15$\\
	$\pm1$ & $0$ & $0$ & $0.42$\\
	$0$ & $\pm1$ & $0$ & $0.07$\\
	$0$ & $0$ & $\pm1$ & $0.025$\\
	$\sim\mathcal{U}(-1, 1)$ & $\sim\mathcal{U}(-1, 1)$ & $\sim\mathcal{U}(-1, 1)$ & 0.1\\
	\hline
	\end{tabular}
\end{table}

To obtain the noisy exteroceptive student observation $o^n_t$ we apply a noise to the noise free exteroceptive teacher observation $o^e_t$. We sample noises for the sampling coordinates $x, y$ and the sampled height values $z$ at the episode, foot and timestep level. Additionally, we select random points on the height sampling pattern and apply a large noise to simulate outliers. For the parametrization of these noises we leverage three modes \{\textit{nominal}, \textit{offset}, \textit{noisy}\} defined in prior work by~\cite{Miki_2022} who designed the nosing method to mimic noise in real exteroceptive sensors.

\subsection{Reward}
For the teacher policy to learn to follow arbitrary commands over arbitrary terrain we use a number of reward terms that are divided into three main categories: (1) \textit{gait}, (2) \textit{command following} and (3) \textit{smoothness}. The full reward function we use is defined as:
\begin{multline*}
	r = (0.25r_{frc} + 0.25r_{vel} + 0.2) \cdot c_r + (r_{air} + 0.1r_{one}) \cdot (1-c_r) \\
	+ 0.2r_{v,xy} + 0.2r_{\omega,z} + 0.05r_{lov} + 0.05r_{fo} \\
	+ 0.05r_{pm} + 0.05r_{po} + 0.025r_{t} + 0.025r_{a}
\end{multline*}
of which the components are explained in the next subsections. 

\subsubsection{Gait rewards}
\begin{figure}[tbp]
	\centerline{\includegraphics[width=\linewidth]{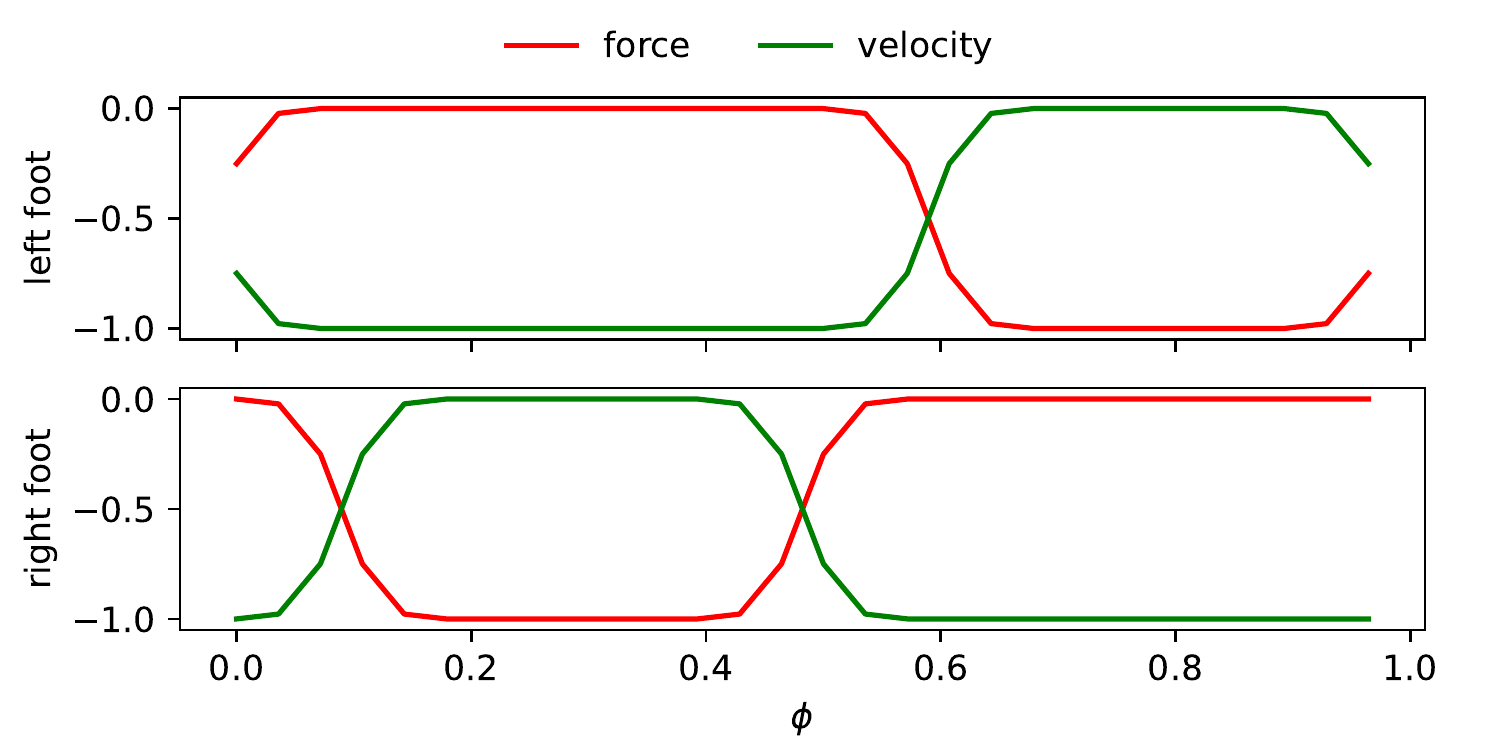}}
	\caption{
		The gait clocks $k_{frc}(\phi)$ and $k_{vel}(\phi)$ for both feet. \
		The stance phases of both feet are offset by $\phi/2$, but overlap slightly, producing a walking gait.
	}
	\label{fig:clock_reward}
\end{figure}

We use a combination of two reward methods to learn a gait. The first is a clock based reward function that oscilates between swing and stance modes in a gait period, as introduced in~\cite{siekmann_godse}. In the swing phase the reward function will penalize foot forces, while in the stance phase foot velocity is penalized. By offsetting the clock functions for the left and right leg a gait can be learned. Figure~\ref{fig:clock_reward} shows the gait clocks.
The foot force reward component $r_{frc}$ is defined as:
\begin{equation*}
	r_{frc} = \tanh(\pi F_l k_{frc,l}) + \tanh(\pi F_r k_{frc,r})
\end{equation*}
where $F_r$ and $F_l$ represent the normalized norm of the foot forces on the respective foot, and $k_{frc,l}$ and $k_{frc,r}$ are the respective foot force gait clocks. The foot velocity reward component $r_{vel}$ is defined as:
\begin{equation*}
	r_{vel} = \tanh(\pi v_l k_{vel,l}) + \tanh(\pi v_r k_{vel,r})
\end{equation*}
where $v_r$ and $v_l$ represent the normalized norm of the foot velocities of the respective foot, and $k_{vel,l}$ and $k_{vel,r}$ are the respective foot velocity gait clocks.

We find that this periodic reward function produces good quality gaits on flat terrain, however performance on rough terrain is less satisfactory. We hypothesize that this is due to the fixed cadence embedded in the gait clocks, restricting the policy to a fixed gait period, and that a more flexible reward function is better suited for rough terrain.

The second is a more flexible reward function that simply rewards foot airtime~\cite{rudin2021parallel} for both feet. The foot airtime reward component $r_{air}$ is defined as:
\begin{equation*}
	r_{air} = \sum_{f=0}^{2} (\mathbf{t}_{air,f} - 0.5) * \vmathbb{1}_{first\ contact,f}
\end{equation*}
where $\mathbf{t}_{air} \in \mathbb{R}^2$ denotes the cumulative airtime during the swing phase of both feet, and $\vmathbb{1}_{first\ contact} \in \mathbb{R}^2$ holds binary values indicating whether the swing phase is ended by contact. To prevent the policy from learning to simply jump, a component $r_{one}$ is added to reward standing on one foot:
\begin{equation*}
	r_{one} = \vmathbb{1}_{single\ contact}
\end{equation*}
where $\vmathbb{1}_{single\ contact}$ is a binary value indicating whether the robot is standing on one foot. We find that this reward function leads to more stable gaits on rough terrain, but convergence is slower. Therefore we start the training process with the clock based reward function until a satisfactory gait has been learned on flat terrain. We then switch to the airtime based reward function by setting the reward curriculum factor $c_r$ from 1 to 0.

\subsubsection{Command following rewards}
We employ similar command following rewards as~\cite{Lee_2020,Miki_2022} with the goal of maximizing velocity in a given direction. The velocity reward component $r_{v,xy}$ is defined as:
\begin{equation*}
	r_{v,xy} = \begin{cases}
		\exp(-2.5 \cdot \left|\left| \mathbf{v}_{xy} \right|\right|^2) & \left|\left| \mathbf{v}_{cmd} \right|\right| = 0 \\
		1 & \mathbf{v}_{cmd} \cdot \mathbf{v}_{xy} \geq 1 \\
		\exp(-2 \cdot (\mathbf{v}_{cmd} \cdot \mathbf{v}_{xy} - 1)^2) & else
	\end{cases}
\end{equation*}
where $\mathbf{v}_{xy} \in \mathbb{R}^2$ represents the linear velocity in the $xy$ plane. The angular velocity reward component $r_{\omega,z}$ is defined as:
\begin{equation*}
	r_{\omega,z} = \begin{cases}
		\exp(-5 \cdot \omega_z^2) & \mathbf{\omega}_{cmd} = 0 \\
		1 & \omega_{cmd} \cdot \omega_z \geq 1 \\
		\exp(-2 \cdot (\omega_{cmd} \cdot \omega_z - 1)^2) & else
	\end{cases}
\end{equation*}
where $\omega_{z}$ represents the pelvis angular velocity. The linear orthogonal velocity offset reward component $r_{lov}$ is defined as:
\begin{equation*}
	r_{lov} = \exp(-5 \cdot \left|\left| \mathbf{v}_{xy} - \mathbf{v}_{cmd} \cdot \mathbf{v}_{xy} \right|\right|)
\end{equation*}
where $\mathbf{v}_{xy} \in \mathbb{R}^2$ again represents pelvis velocity in the $xy$ plane. It is intended to penalize linear velocities orthogonal to the commanded velocity.

\subsubsection{Smoothness rewards}
The foot orientation reward component $r_{fo}$ is defined as:
\begin{equation*}
	r_{fo} = \exp(-1.5 \cdot (\mathbf{\hat{z}} \cdot \boldsymbol{\psi}_{lf} + \mathbf{\hat{z}} \cdot \boldsymbol{\psi}_{rf})) \cdot (1-c_t) + c_t
\end{equation*}
where $\mathbf{\hat{z}} \in \mathbb{R}^3$ represents the unit vector in the $z$ direction and $\boldsymbol{\psi}_{lf}, \boldsymbol{\psi}_{rf}  \in \mathbb{R}^3$ represent the foot orientation vectors pointing along the length of both feet. This reward component encourages the policy to keep the feet flat on the ground, but in any planar direction. Additionally, the reward is gradually shifted to a constant reward component as the curriculum is ramped up to allow the policy to adapt to terrains where a non flat position might be more beneficial. The pelvis motion reward component $r_{pm}$ is defined as:
\begin{equation*}
	r_{pm} = \exp(-(v_z^2 + \omega_y^2 + \omega_x^2))
\end{equation*}
and is intended to penalize pelvis motions in directions not part of the command. The pelvis orientation reward component $r_{po}$ is defined as:
\begin{equation*}
	r_{po} = \exp(-3 \cdot (\left| \psi_x \right| + \left| \psi_y \right|))
\end{equation*}
where $\psi_x$ and $\psi_y$ represent pelvis orientation and is intended to encourage the policy to keep the pelvis level. The torque reward component $r_{t}$ is defined as:
\begin{equation*}
	r_{t} =  \exp(-0.02 \cdot \overline{\left| \boldsymbol{\tau} \right|})
\end{equation*}
where $\boldsymbol{\tau}$ represents the torque vector exerted by the actuators, with the aim to reduce energy consumption. The action reward component $r_{a}$ is defined as:
\begin{equation*}
	r_{a} = \exp(-5 \cdot \overline{\left| \mathbf{a}_t - \mathbf{a}_{t-1} \right|})
\end{equation*}
where $\mathbf{a}_t$ represents the action vector at time $t$ and is intended to penalize large changes in the action vector in order to improve smoothness and stability.

\section{Experimental Results}
\label{sec:experimental_results}
\subsection{Simulation}
Training and experimentation are performed in simulation on the Cassie robot~\cite{agility_robotics}. We use the Mujoco~\cite{mujoco} physics simulator with the \textit{cassie-mujoco-sim} environment~\cite{cassieMujocoSim}. 

\subsection{Training}
For teacher policy training the recurrent PPO algorithm from StableBaselines3 (SB3)~\cite{stable-baselines3} is used. In order to achieve a 7 times speedup in terms of timesteps per second we modified SB3 to use batches of whole sequences. The hyperparameters for PPO are listed in Table~\ref{tab:ppo_hyperparams}. Teacher policy training takes around 36 hours for $60 \times 10^6$ timesteps on a single 12-core V100 node.

The student policy is trained on a dataset of $10^6$ timesteps sampled from the teacher policy, with the hyperparameters listed in Table~\ref{tab:student_hyperparams}. The student policy is implemented in PyTorch~\cite{PyTorch}, and training takes around 1 hour on a high end laptop.

\begin{table}[t]
	\renewcommand{\arraystretch}{1.3}
	\centering
	\caption{Policy Training Hyperparameters}
	\subfloat[Teacher policy]{
		\label{tab:ppo_hyperparams}
		\begin{tabular}{l||l}
		\hline
		\bfseries Parameter & \bfseries Value\\
		\hline\hline
		max episode length & 300 \\
		rollout buffer size & 50000\\
		learning rate & $10^{-4}$ \\
		batch size & 8 seq. \\
		epochs & 5 \\
		clip range & 0.2 \\
		GAE Lambda & 0.95 \\
		gamma & 0.99 \\
		\hline
	\end{tabular}
	}
	\hfill
	\subfloat[Student policy]{
		\label{tab:student_hyperparams}
		\begin{tabular}{l||l}
		\hline
		\bfseries Parameter & \bfseries Value\\
		\hline\hline
		max episode length & 300 \\
		learning rate & $0.001$ \\
		batch size & 12 seq. \\
		epochs & 100 \\
		optimizer & Adam \\
		\hline
	\end{tabular}
	}
\end{table}

In order to compare and quantify the performance of our exteroceptive student policy we train a baseline policy $\pi^b$. This baseline policy observes proprioception $\mathbf{o}_t^p$ only and has the same architecture as the teacher policy but without the exteroceptive encoder, similar to the current state-of-the-art~\cite{Siekmann-RSS-21,siekmann_godse}. The baseline policy is trained on the same terrain and curriculum as the exteroceptive student policy.

\subsection{Experiments}
We conduct a number of simulation based experiments to evaluate the performance of our exteroceptive policy against the proprioceptive baseline policy. In all experiments our policy has access to \textit{nominal} noise exteroception, and is commanded to walk forward, unless otherwise mentioned. For all results 100 episodes were attempted and the results averaged. 
We include a supplementary video providing a visual representation of the experiments. \footnote{\url{https://youtu.be/B3Qr-7ZZHZQ}}

\subsubsection{Maximum speed over various terrains}
\begin{figure*}[htbp]
	\centerline{\includegraphics[width=\linewidth]{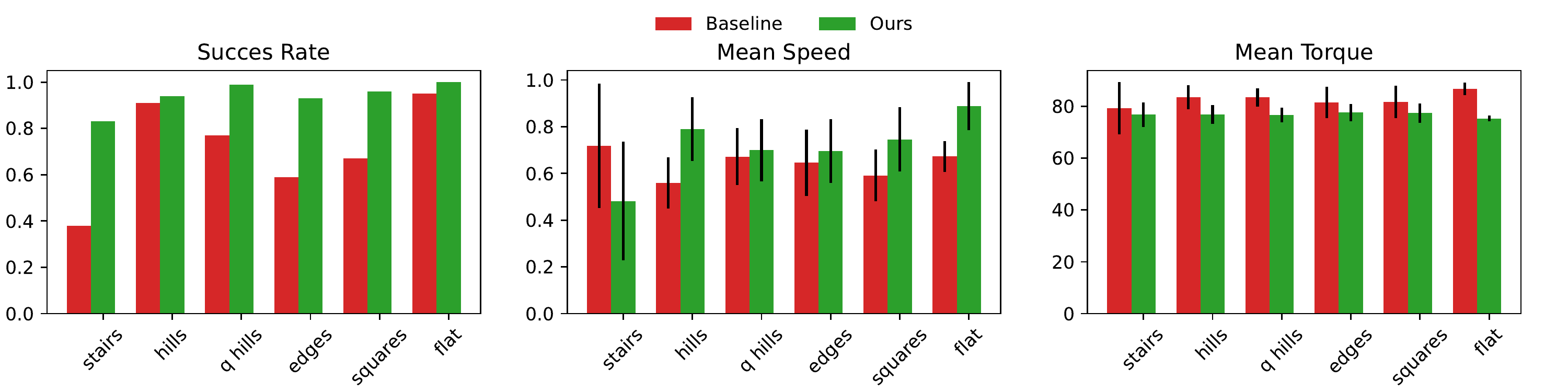}}
	\caption{Policy performance metrics for the five different terrain modes as well as flat terrain. 
	Error bars denote the standard deviation of the mean.
	Our exteroceptive policy achieves higher success rates and speeds as it can take more decisive actions, thanks to its ability to gather information about the environment in advance.
	}
	\label{fig:benchmark_bar_plot}
\end{figure*}

\begin{figure}[htbp]
	\centerline{\includegraphics[width=\linewidth]{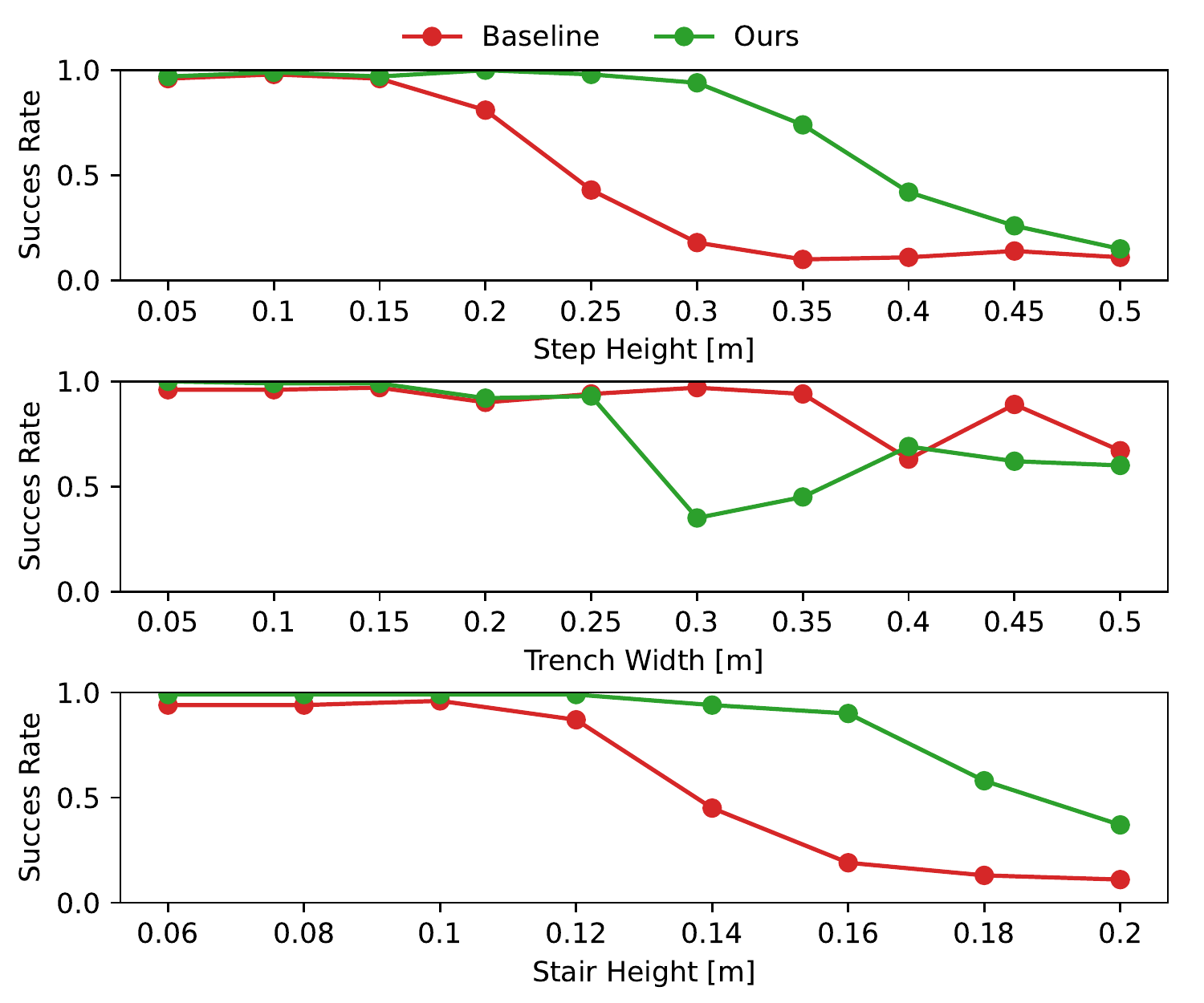}}
	\caption{Success rate metrics for stepping over a single step (top), a trench (middle) and walking downstairs (bottom). In all cases the exteroceptive policy is able to traverse more challenging terrains than the proprioceptive baseline policy.}
	\label{fig:benchmark_step_trench_stair}
\end{figure}
We record mean speed, average actuator torque and the success rate for all terrains in the training curriculum. A success is defined as the episode completing at 300 timesteps without the robot falling over. The results are shown in Figure~\ref{fig:benchmark_bar_plot}. 
Our policy outperforms the baseline policy on all terrains in terms of success rate, with near perfect scores on all terrains but stairs. Although the proprioceptive baseline is able to achieve a near 100\% success rate on hills and flat terrain, its performance is worse on quantized hills, edges and squares. The largest outperformance is found in the stairs terrain, and upon further investigation we find that most failures occur during the descent of the staircase. This underperformance of the baseline policy is likely due to the large vertical change in the stairs, quantized hills, edges and squares terrains, which the robot is unable to anticipate due to the lack of exteroceptive information. To investigate this further we conduct a more detailed analysis of the staircase terrain in the next experiments.

Our exteroceptive policy is able to achieve higher speeds than the baseline policy on all terrains but stairs. This is due to the fact that the exteroceptive policy is able to take more decisive actions, as it can gather information about the environment in advance. The baseline policy is forced to resort to more conservative gaits, at slower speeds and lifting feet higher. 
The slow speed on stairs for our exteroceptive policy is caused by a combination of caution and the policy veering slightly off course during stair descent. This behaviour emerged from the training process and we hypothesize that, although undesirable, it is beneficial for the policy success, as it effectively lengthens the run of the stairs, making it easier to find a suitable foothold.

Applied torque is lower for our policy than the baseline policy on all terrains, indicating that less energy is consumed despite walking at higher speeds. Our exteroceptive policy is therefore more energy efficient. However, we did not investigate whether the difference is large enough to counteract the added computational cost of the exteroceptive encoder.

\subsubsection{Success rate over a step}
We command both policies to go forward over flat terrain with a 1 meter wide step of varying heights at 1 meter from the starting position and record the success rate. The results are shown in Figure~\ref{fig:benchmark_step_trench_stair}. Our policy outperforms the baseline policy, reliably traversing over steps up to 30~cm in height, while the baseline policy starts to fail at 20~cm. This experiment clearly shows the advantage of exteroceptive information in the case of a step, as the policy can anticipate the step and take a more reliable action.

\subsubsection{Success rate over a trench}
We command both policies to go forward over a 50~cm deep trench of varying widths to gauge the effects of exteroception on the policy's ability to avoid dangerous areas. The results are shown in Figure~\ref{fig:benchmark_step_trench_stair}. 
Surprisingly the proprioceptive baseline is able to outperform our exteroceptive policy. We find that the exteroceptive policy does not avoid the trench, but tries to step inside. We hypothesize this is caused by the policy not encountering terrains with dangerous areas during training.

\subsubsection{Success rate for stair descent}
\begin{figure*}
	\centering
	\subfloat{\includegraphics[trim=1100 500 1100 500, clip, width=0.19\linewidth]{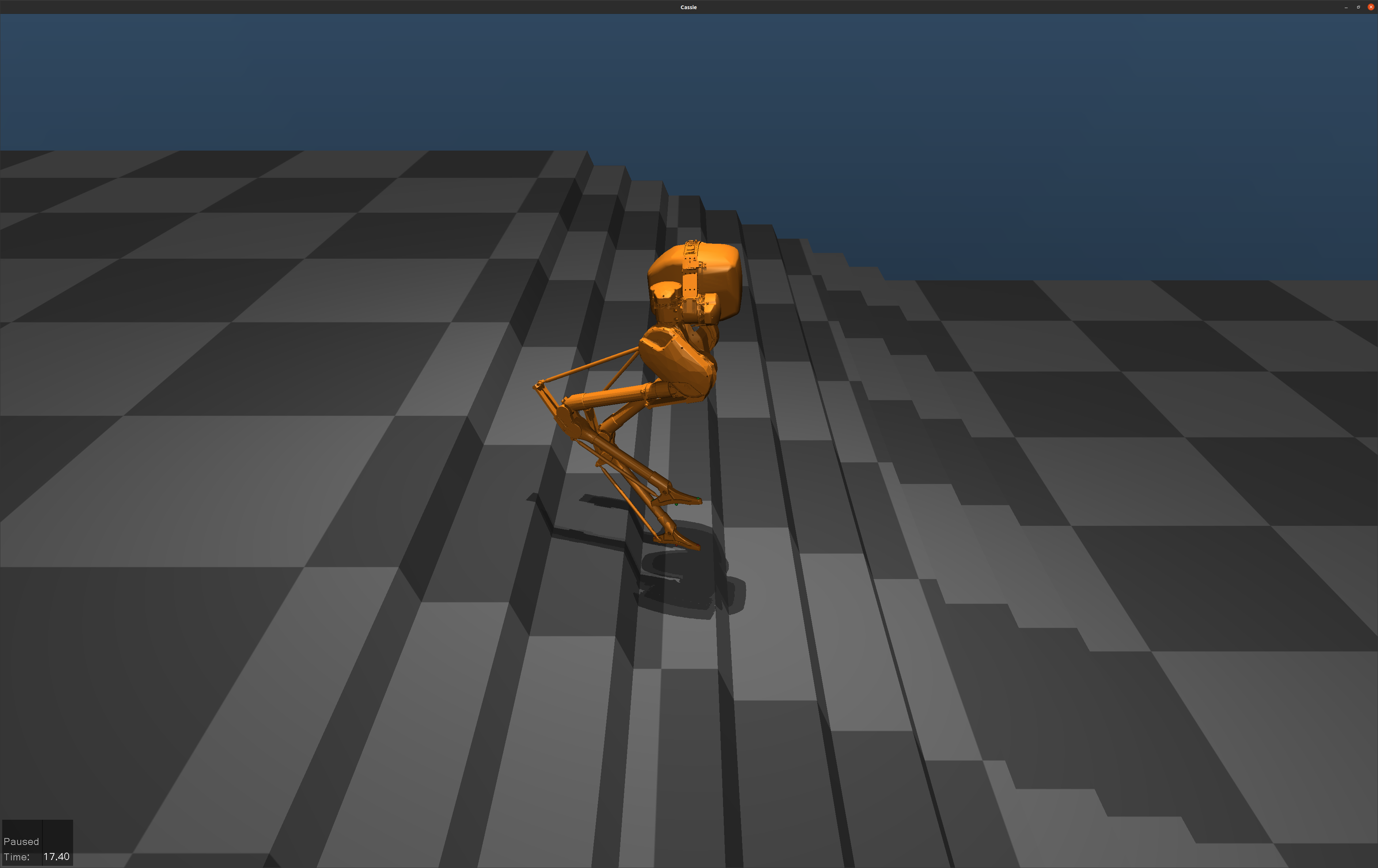}}
	\hspace{1px}
	\subfloat{\includegraphics[trim=1100 500 1100 500, clip, width=0.19\linewidth]{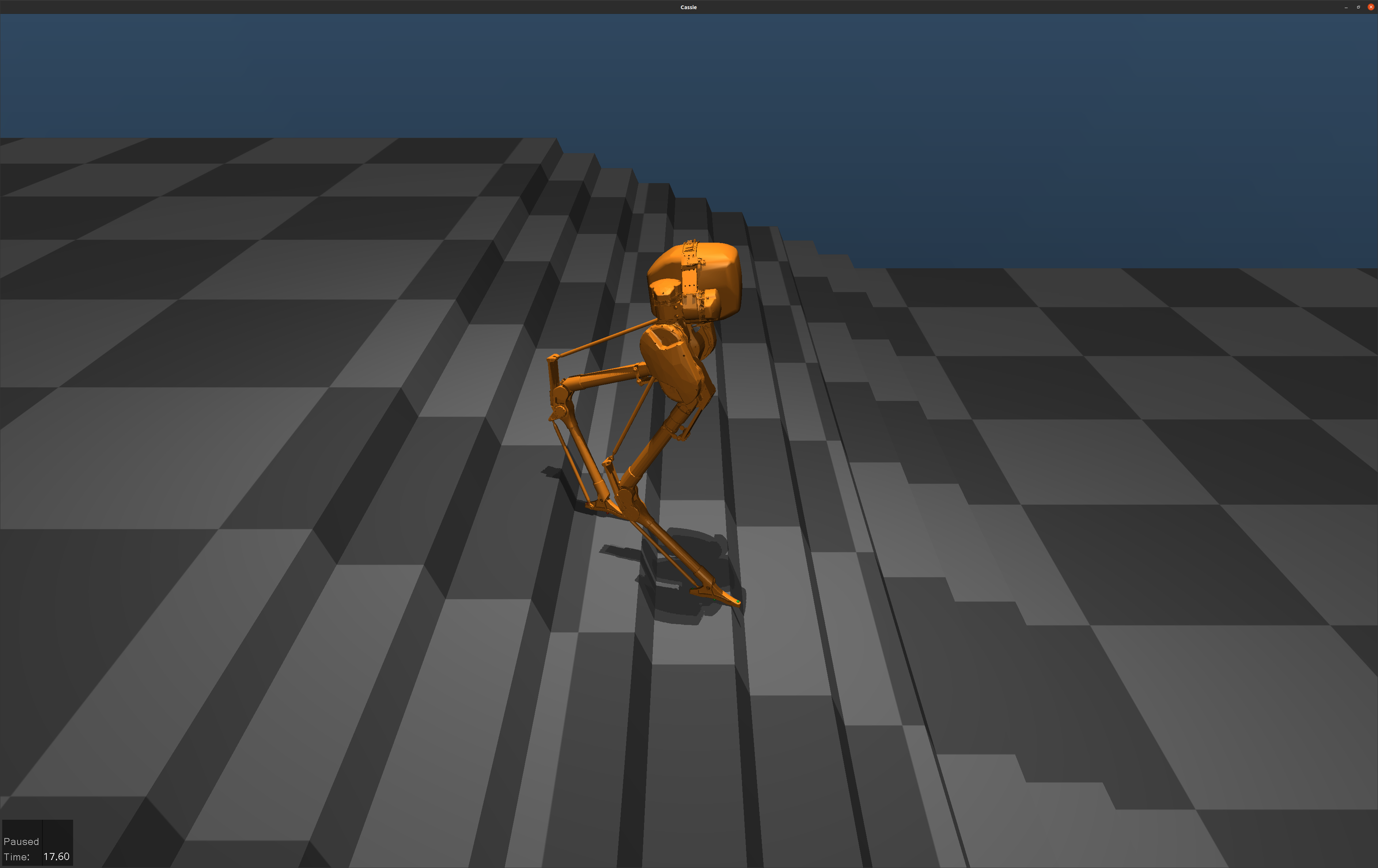}}
	\hspace{1px}
	\subfloat{\includegraphics[trim=1100 500 1100 500, clip, width=0.19\linewidth]{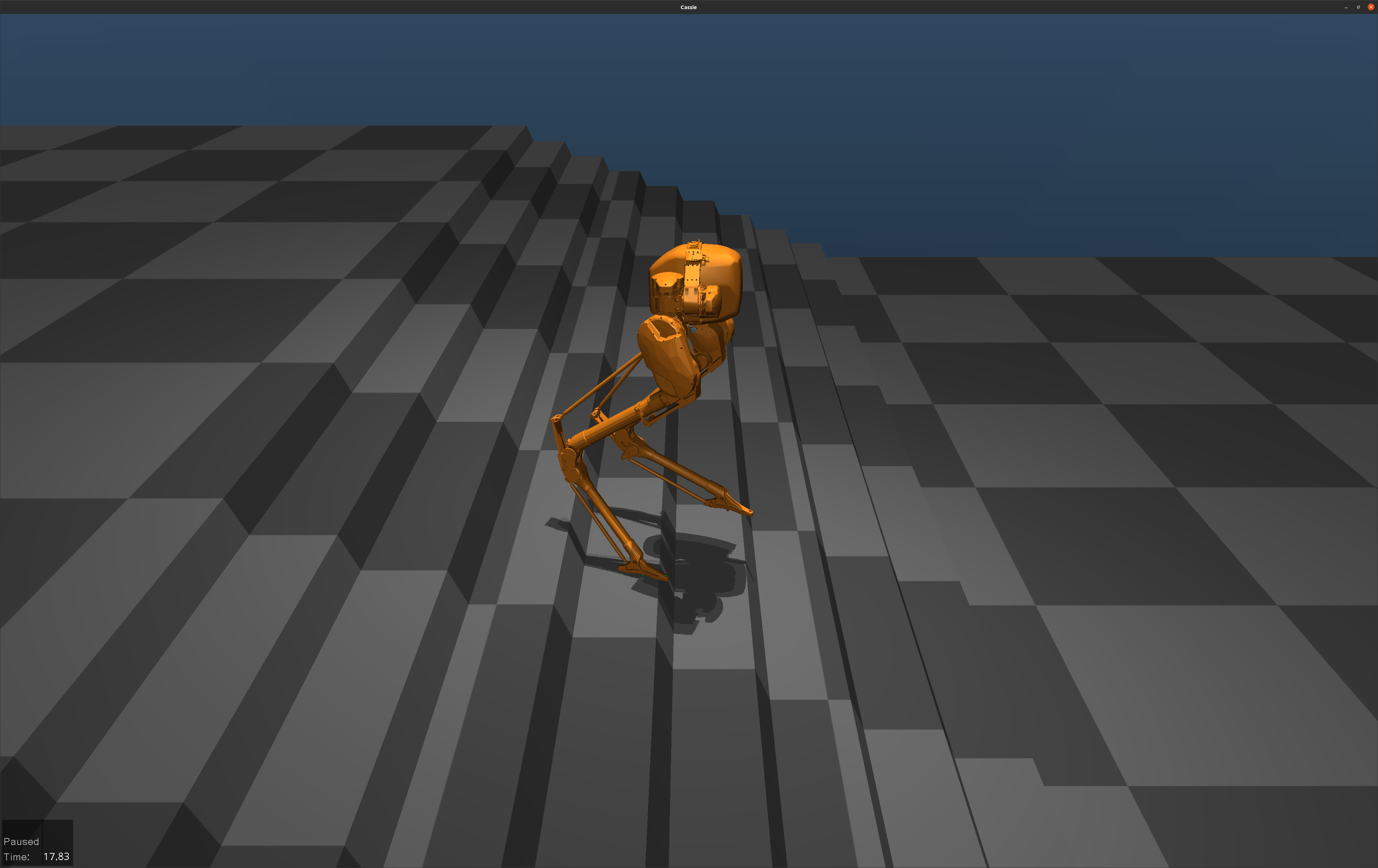}}
	\hspace{1px}
	\subfloat{\includegraphics[trim=1100 500 1100 500, clip, width=0.19\linewidth]{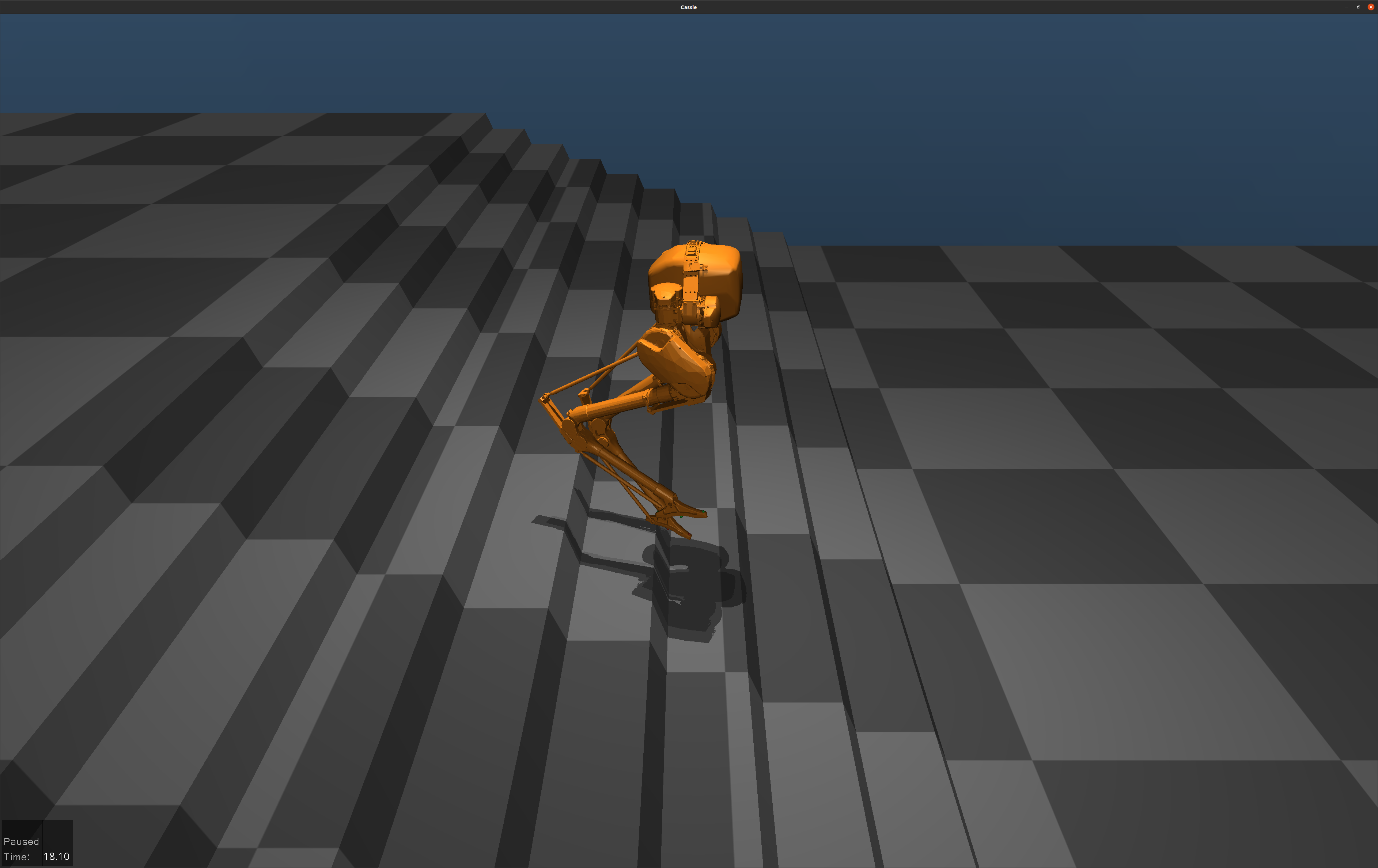}}
	\hspace{1px}
	\subfloat{\includegraphics[trim=1100 500 1100 500, clip, width=0.19\linewidth]{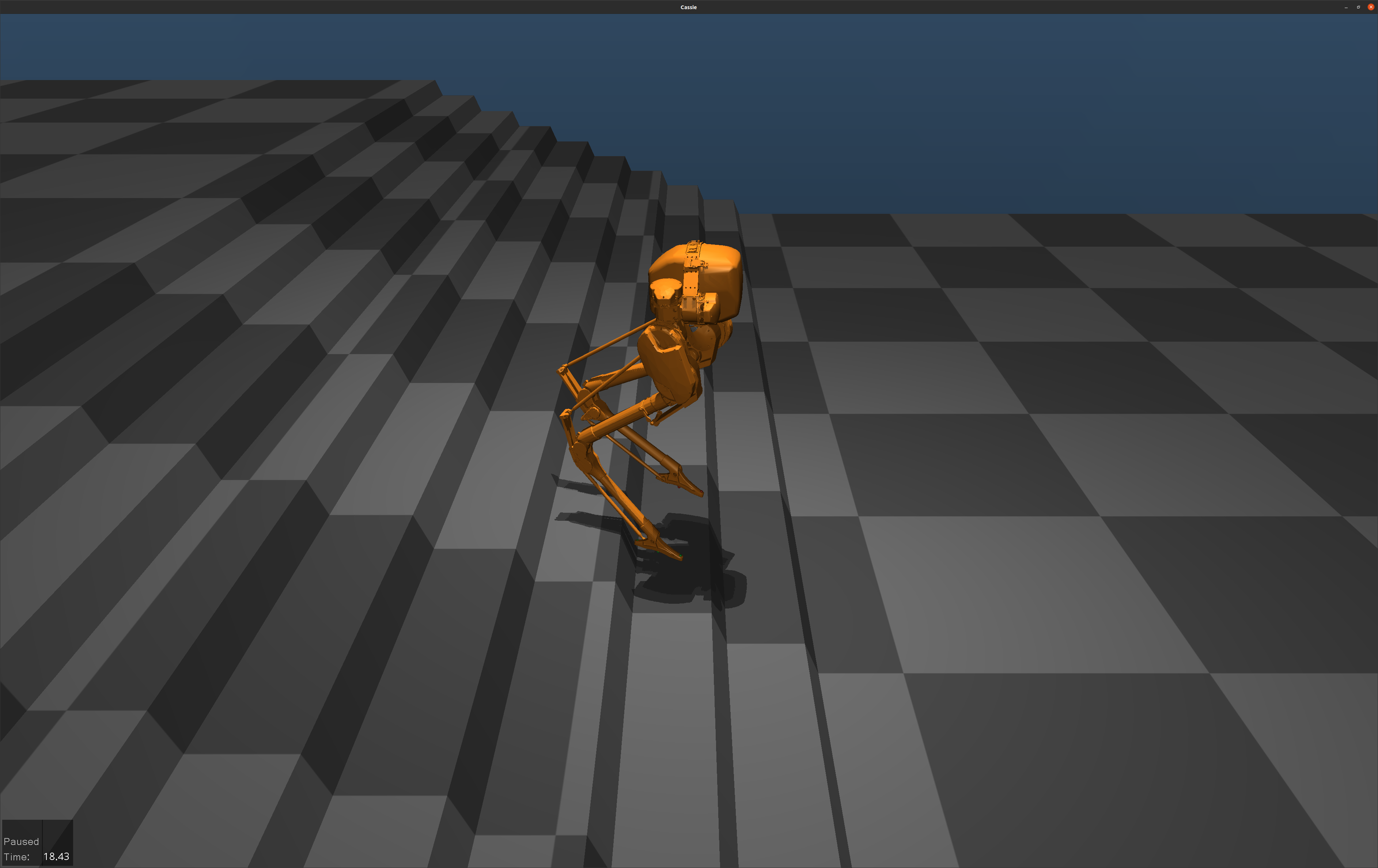}}
	\vspace{-0.2px}
	\subfloat{\includegraphics[trim=330 240 330 400, clip, width=0.19\linewidth]{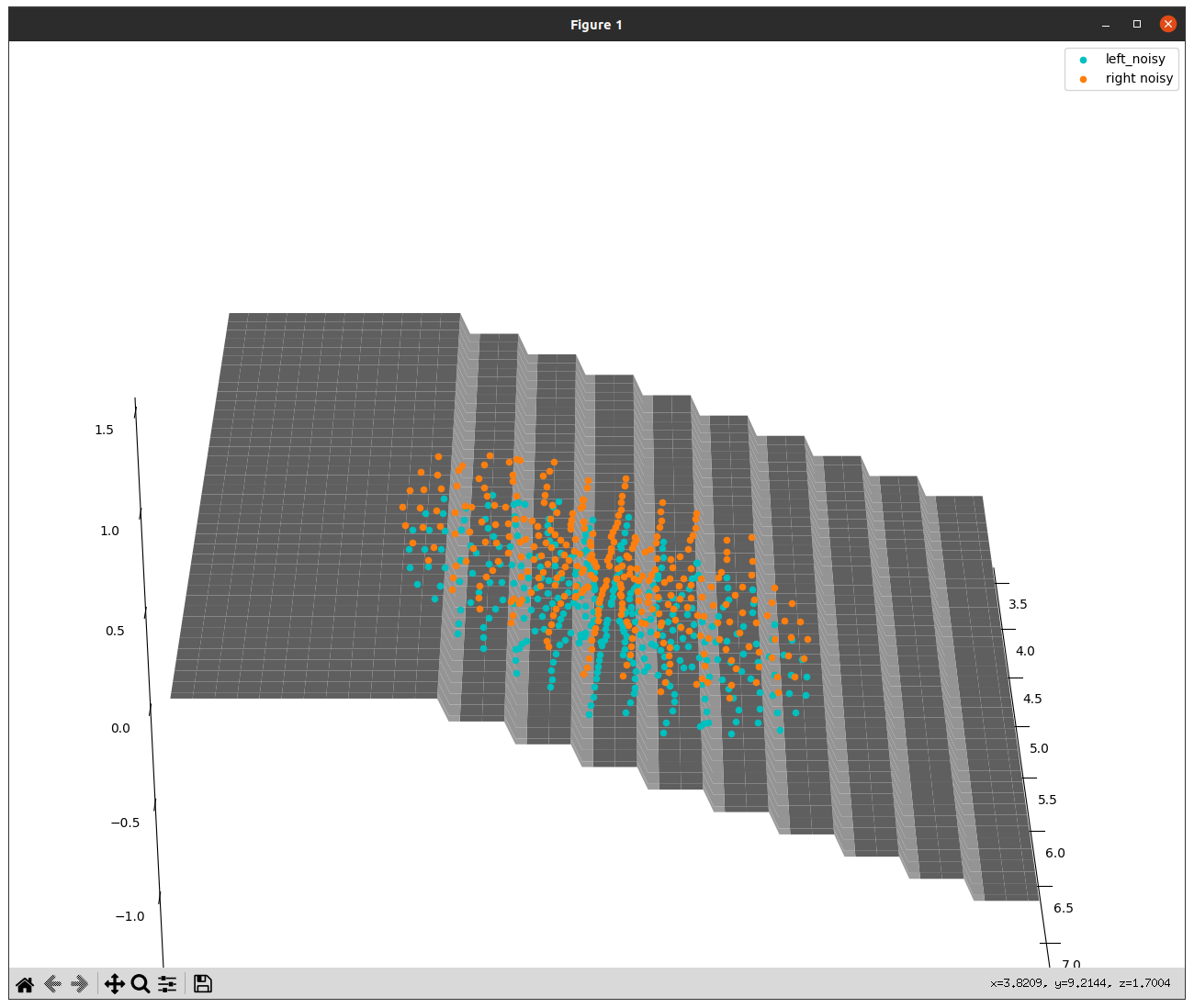}}
	\hspace{1px}
	\subfloat{\includegraphics[trim=330 240 330 400, clip, width=0.19\linewidth]{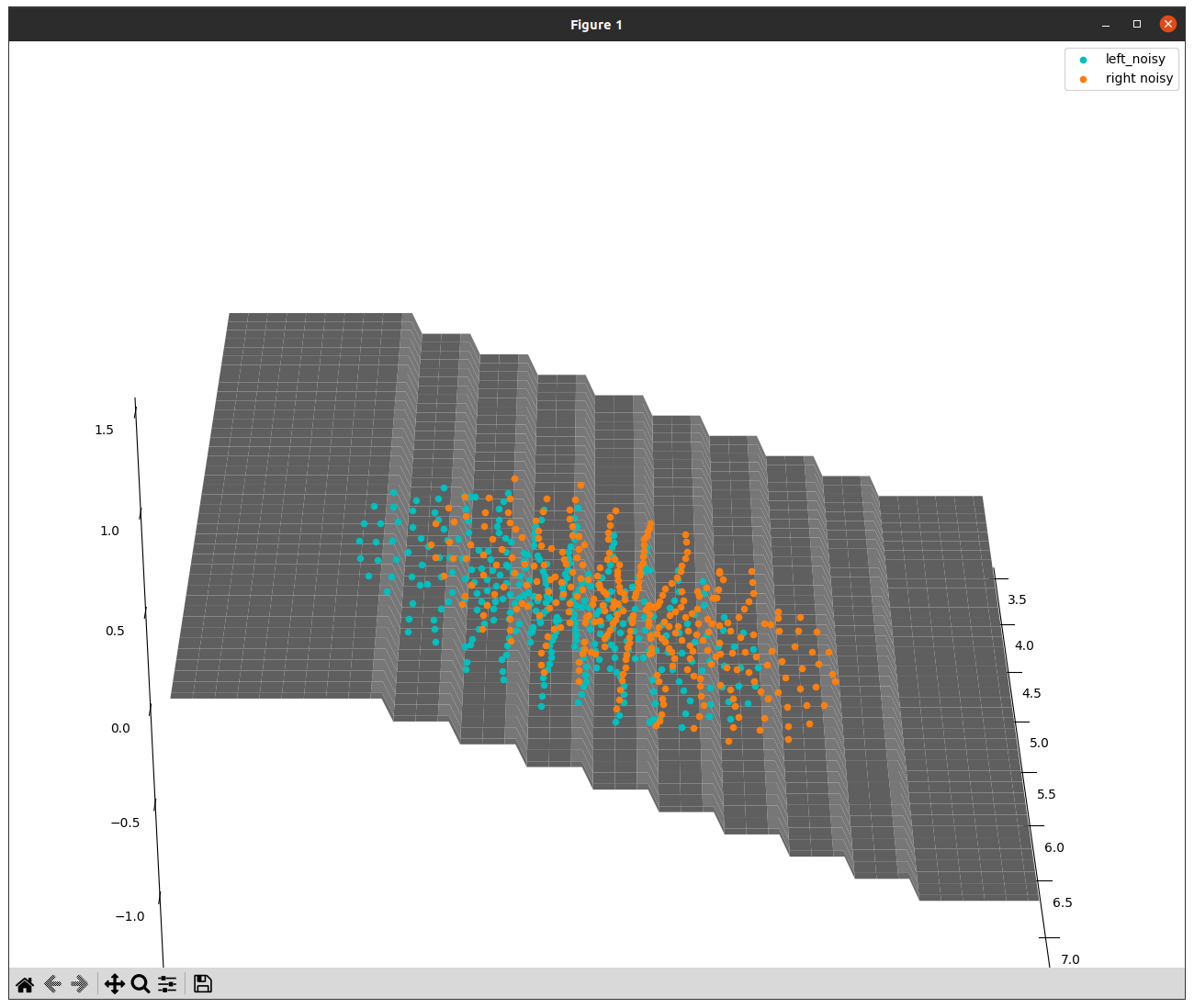}}
	\hspace{1px}
	\subfloat{\includegraphics[trim=330 240 330 400, clip, width=0.19\linewidth]{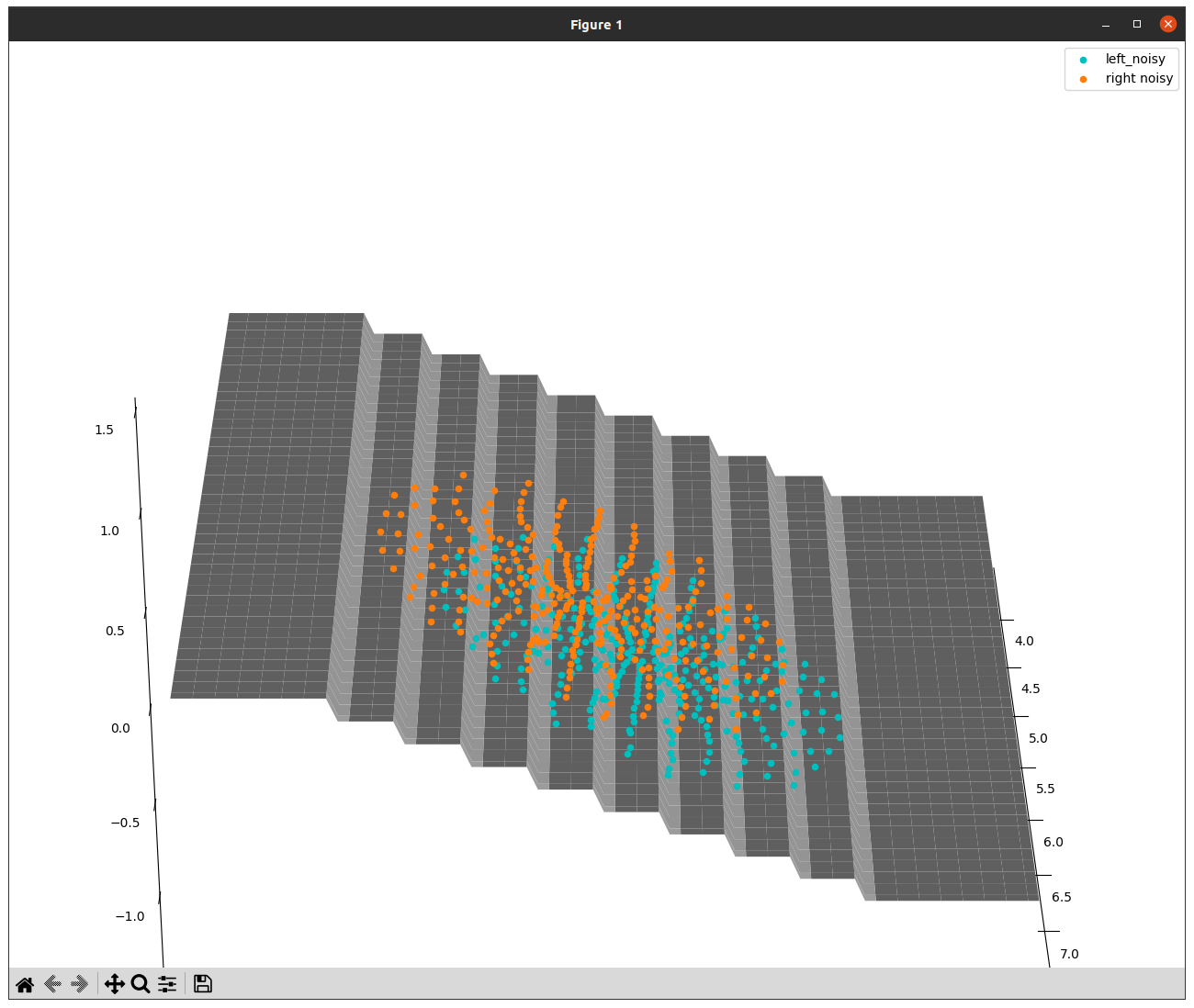}}
	\hspace{1px}
	\subfloat{\includegraphics[trim=330 240 330 400, clip, width=0.19\linewidth]{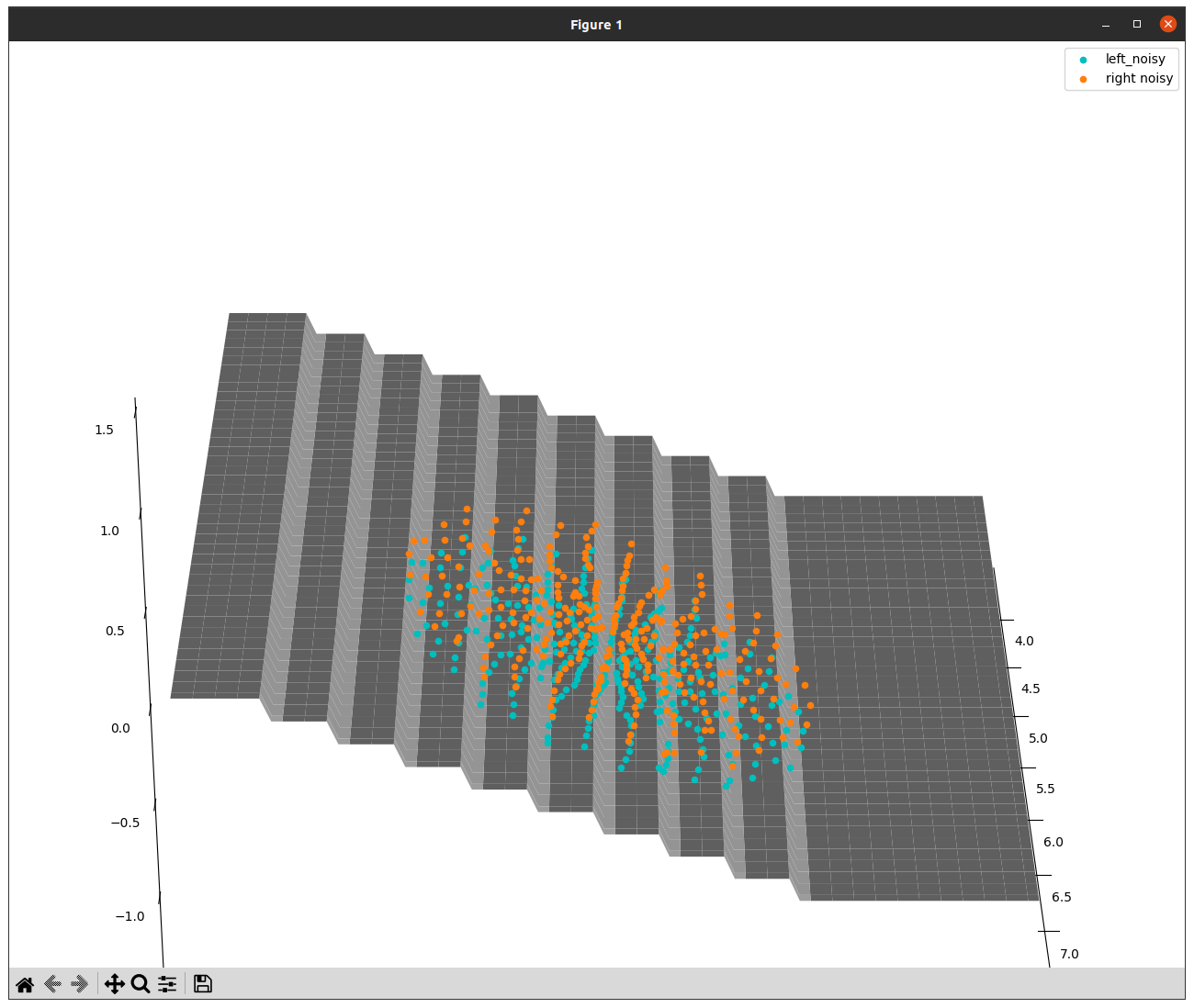}}
	\hspace{1px}
	\subfloat{\includegraphics[trim=330 240 330 400, clip, width=0.19\linewidth]{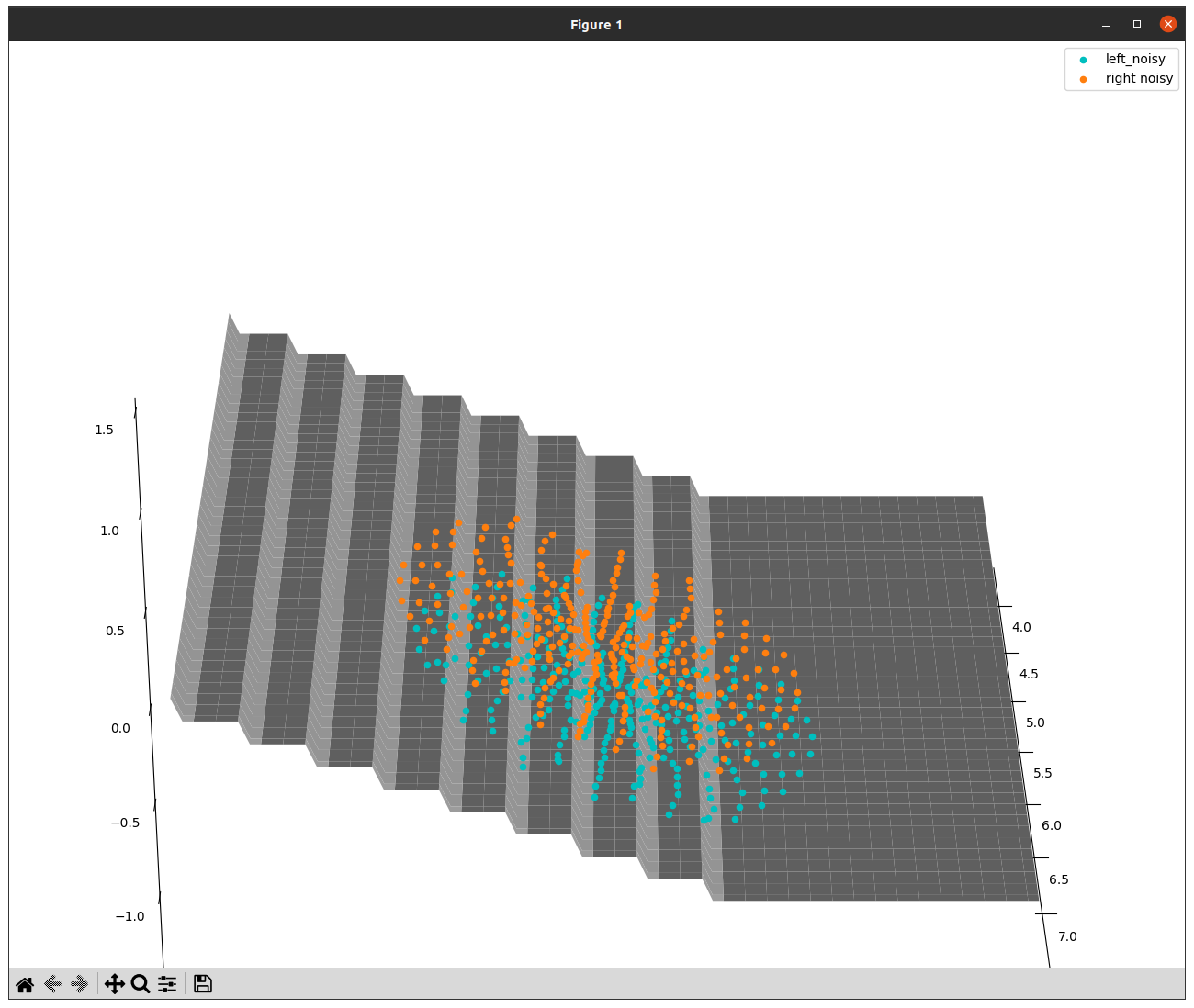}}

	\vspace{2px}

	\subfloat{\includegraphics[trim=1100 500 1100 500, clip, width=0.19\linewidth]{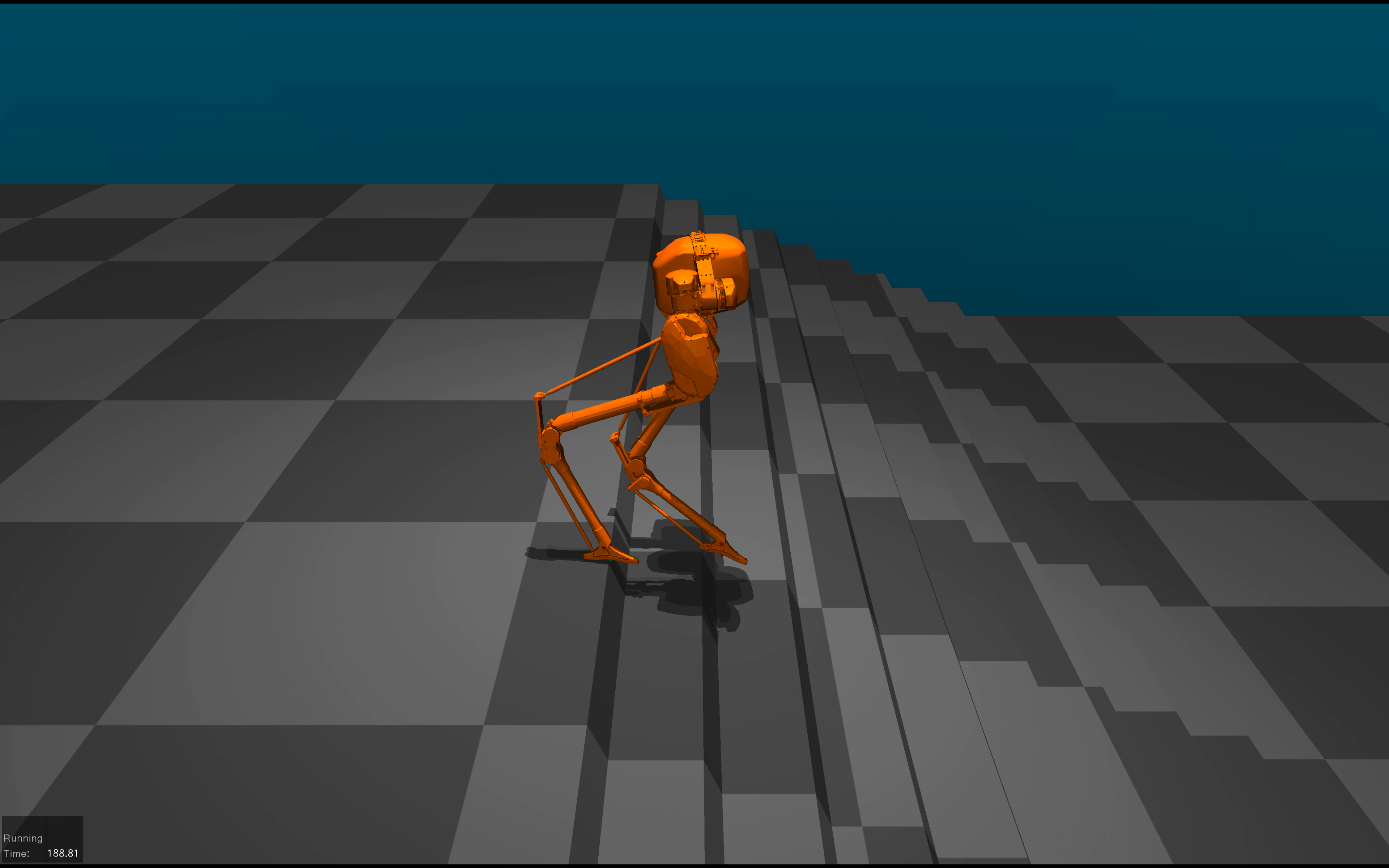}}
	\hspace{1px}
	\subfloat{\includegraphics[trim=1100 500 1100 500, clip, width=0.19\linewidth]{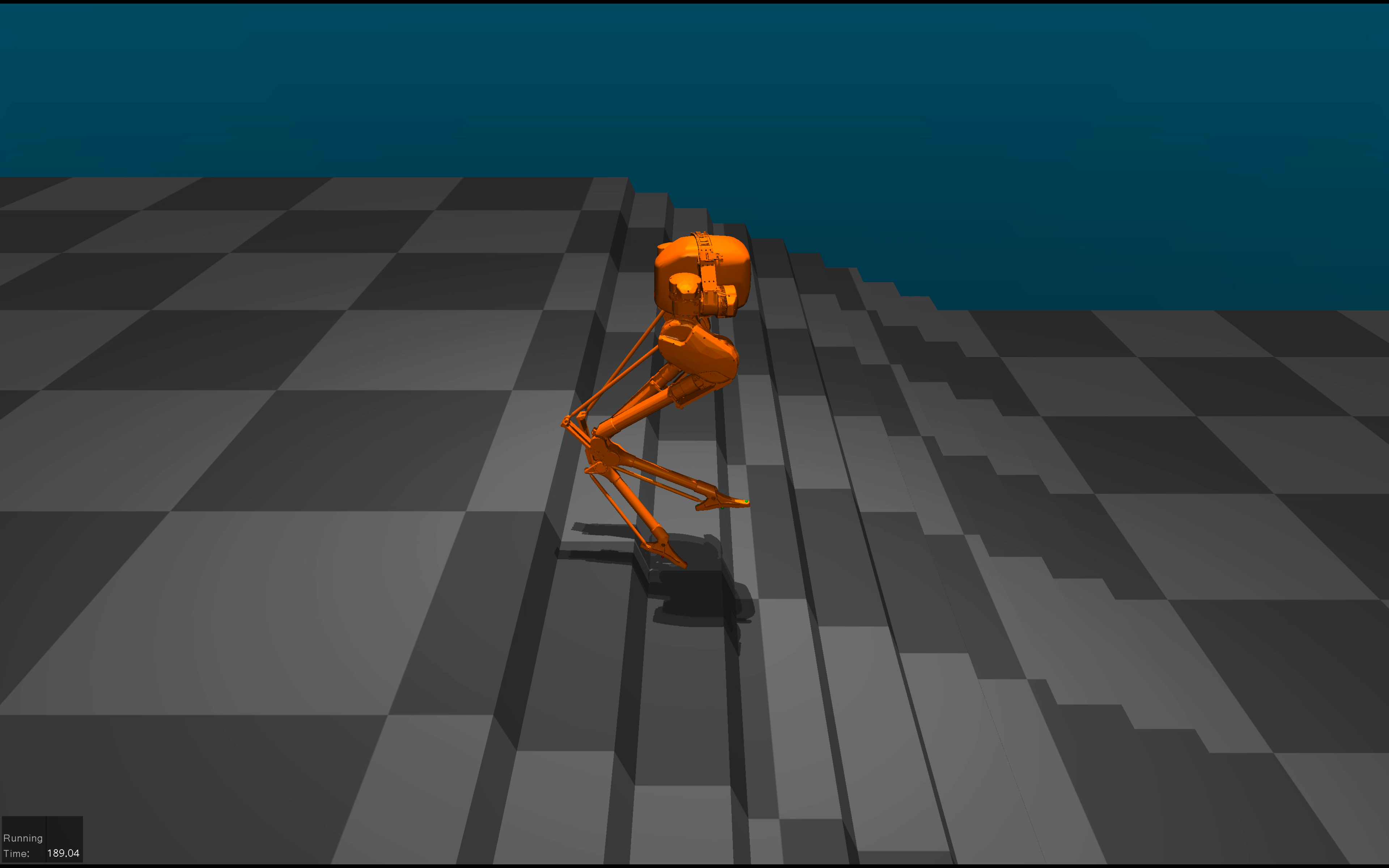}}
	\hspace{1px}
	\subfloat{\includegraphics[trim=1100 500 1100 500, clip, width=0.19\linewidth]{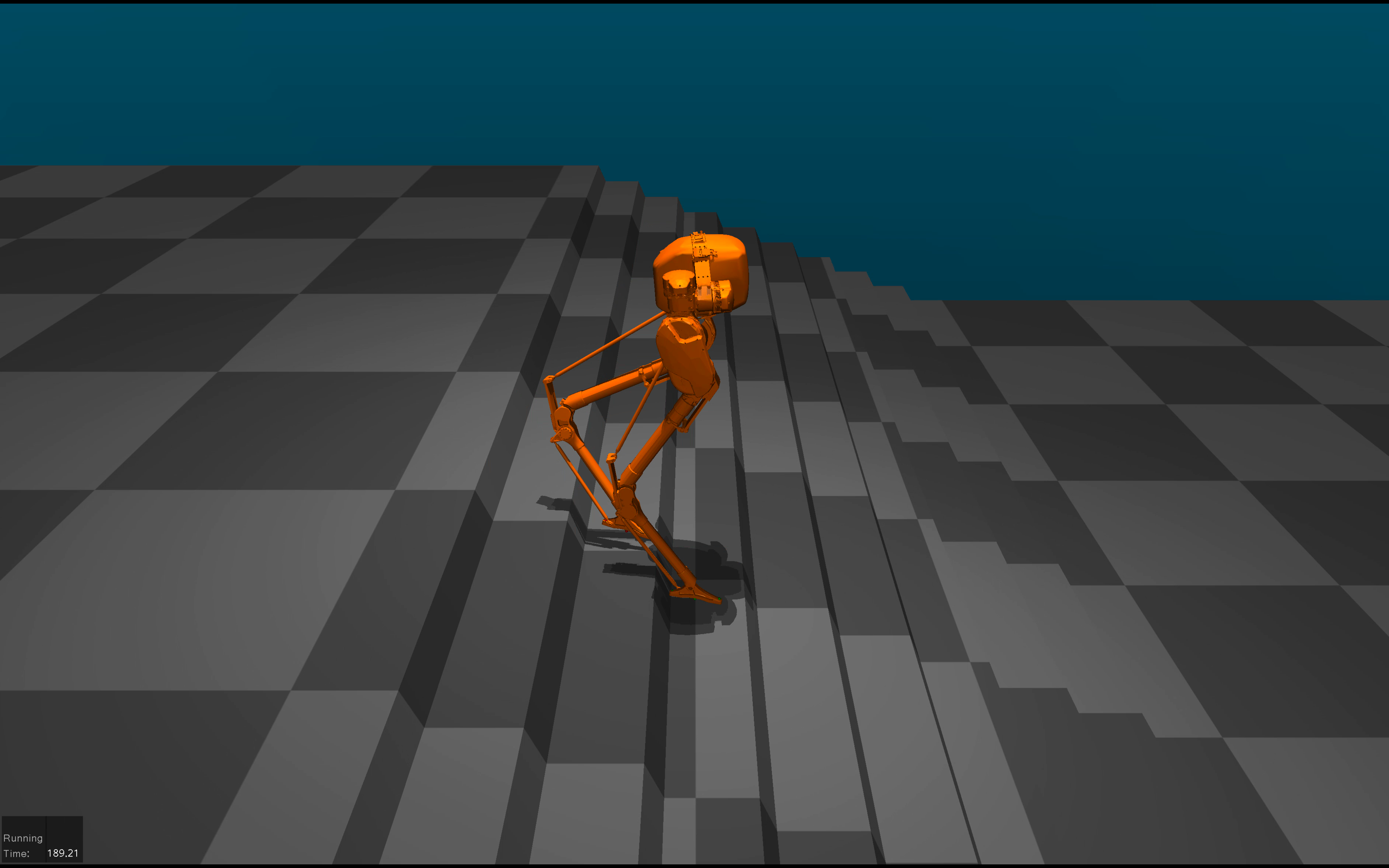}}
	\hspace{1px}
	\subfloat{\includegraphics[trim=1100 500 1100 500, clip, width=0.19\linewidth]{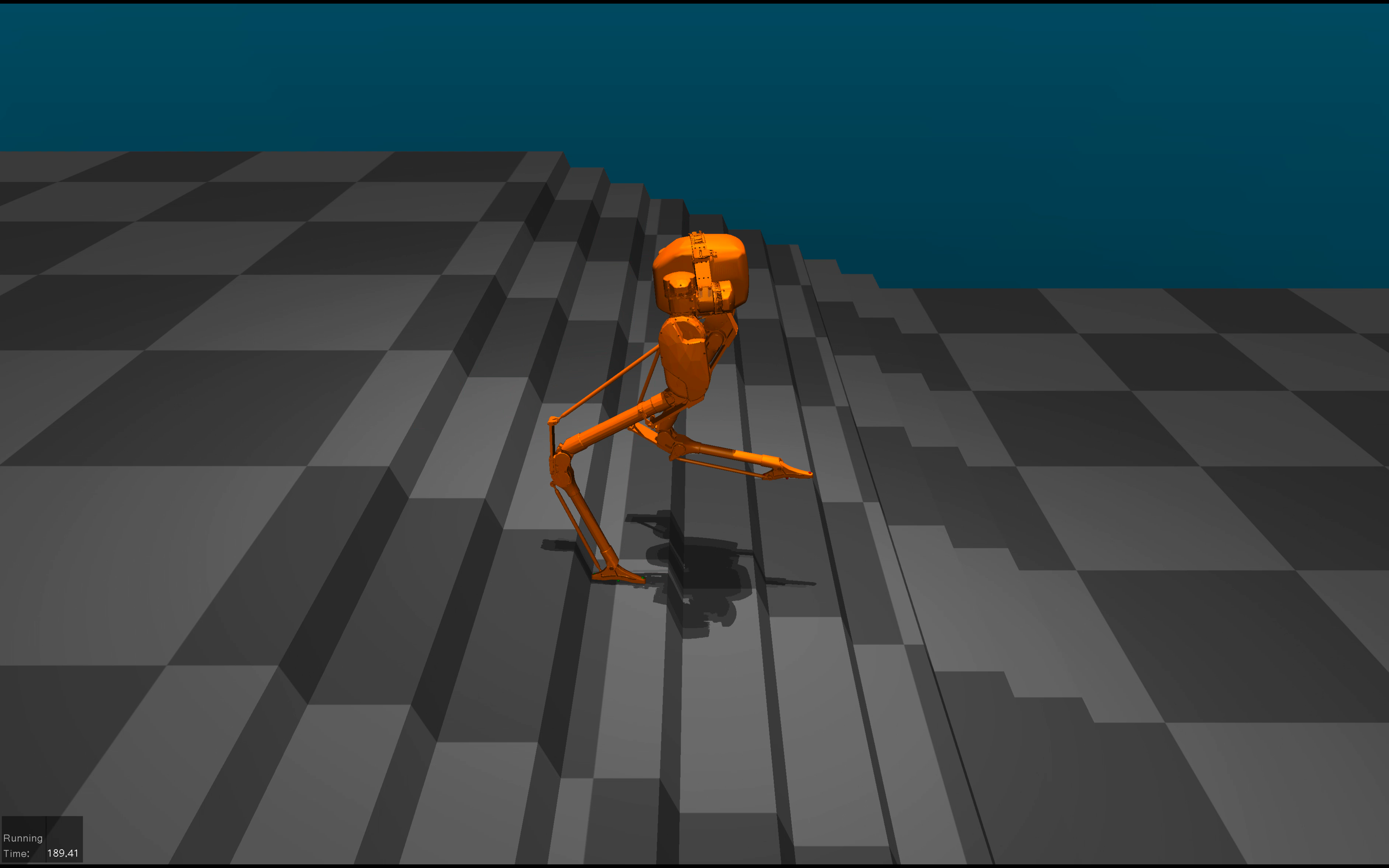}}
	\hspace{1px}
	\subfloat{\includegraphics[trim=1100 500 1100 500, clip, width=0.19\linewidth]{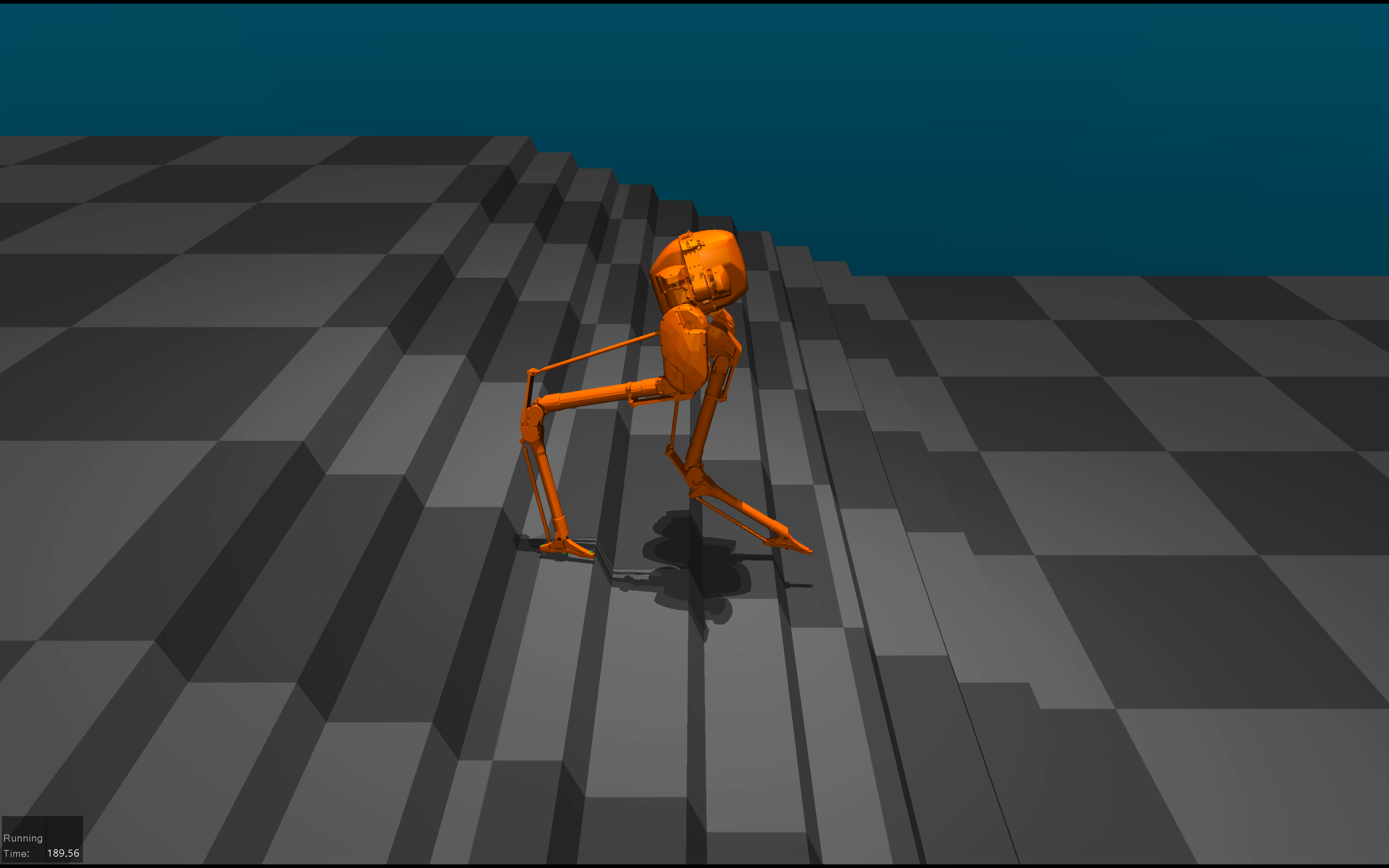}}

	\caption{The top two rows shows our exteroceptive policy successfully walking down a staircase of ten 12~cm steps, with the second row showing the noisy exteroceptive inputs at the same timesteps. The bottom row shows the proprioceptive baseline policy attempting to walk down the same staircase, with less success.}
	\label{fig:benchmark_stair_frames}
\end{figure*}

As can be seen from the success rate results in Figure~\ref{fig:benchmark_bar_plot}, the stair terrain is most difficult for both policies. Specifically, most failures occur when the robot is descending stairs. We believe this is due to Cassies morphology since the steep backwards incline of the legs make them interfere with the stairs. Effectively causing the useable portion of a stair run to be shorter and requiring more precision in foot placement. To investigate whether exteroception is beneficial in this environment we command both policies to go down staircases of 10 equal steps and vary the step heights while recording the success rates. All staircases use the same run of 35~cm. The results are shown in Figure~\ref{fig:benchmark_step_trench_stair}.
Our exteroceptive policy is able to walk down stairs with step heights up to 16~cm with near 100\% success rate, while the baseline policy starts failing at 12~cm. This clearly shows the advantage of exteroception in this environment. Figure~\ref{fig:benchmark_stair_frames} shows both policies traversing down stairs of 12~cm step height.

\begin{figure*}[tbp] 
	\centerline{\includegraphics[width=\linewidth]{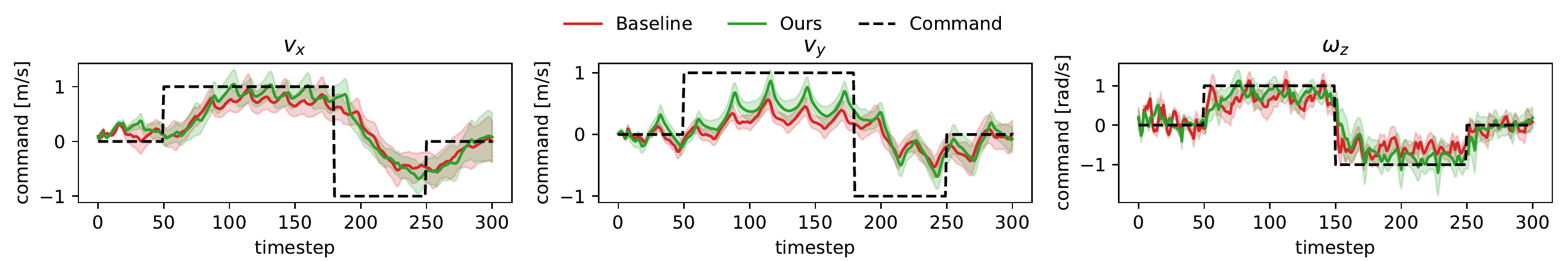}}
	\caption{
			The policy is commanded to walk in all directions over the squares terrain, and the commanded and actual linear and angular velocities are recorded.
			All plots represent the mean velocity of 100 independent trials. The shaded area denotes the standard deviation of the mean.
			}
			\label{fig:command_following}
		\end{figure*}

\subsubsection{Command Following}
To gauge the capability of the policy to locomote over irregular terrain according to a user specified command we command the policy to walk in all directions over the squares terrain and record the speed. We choose the squares terrain for this experiment for its high density of changes in terrain height. Figure~\ref{fig:command_following} shows the commanded linear and angular velocities and the policies responses to them. The policies are able to follow the commands in the $x$ direction over rough terrain accurately, albeit with a delay. Similar results apply for speed achieved in the $y$ direction, however the maximum speed is lower than commanded. We believe this is due to the low range of motion available in the roll actuators of the robot hips, requiring smaller steps and limiting velocity. We found that results are similar on flat terrain, confirming that terrain is not the limiting factor. Lastly, both policies are able to accurately follow the commanded angular velocity over the squares terrain. These results clearly show that the policies have learned to locomote over irregular terrain according to a user specified command, with our exteroceptive policy slightly outperforming in terms of speed.

\subsubsection{Dealing with spurious exteroceptive inputs}
\begin{figure}[tbp]
	\centerline{
		\includegraphics[trim=1300 900 1300 1000, clip, width=\linewidth]{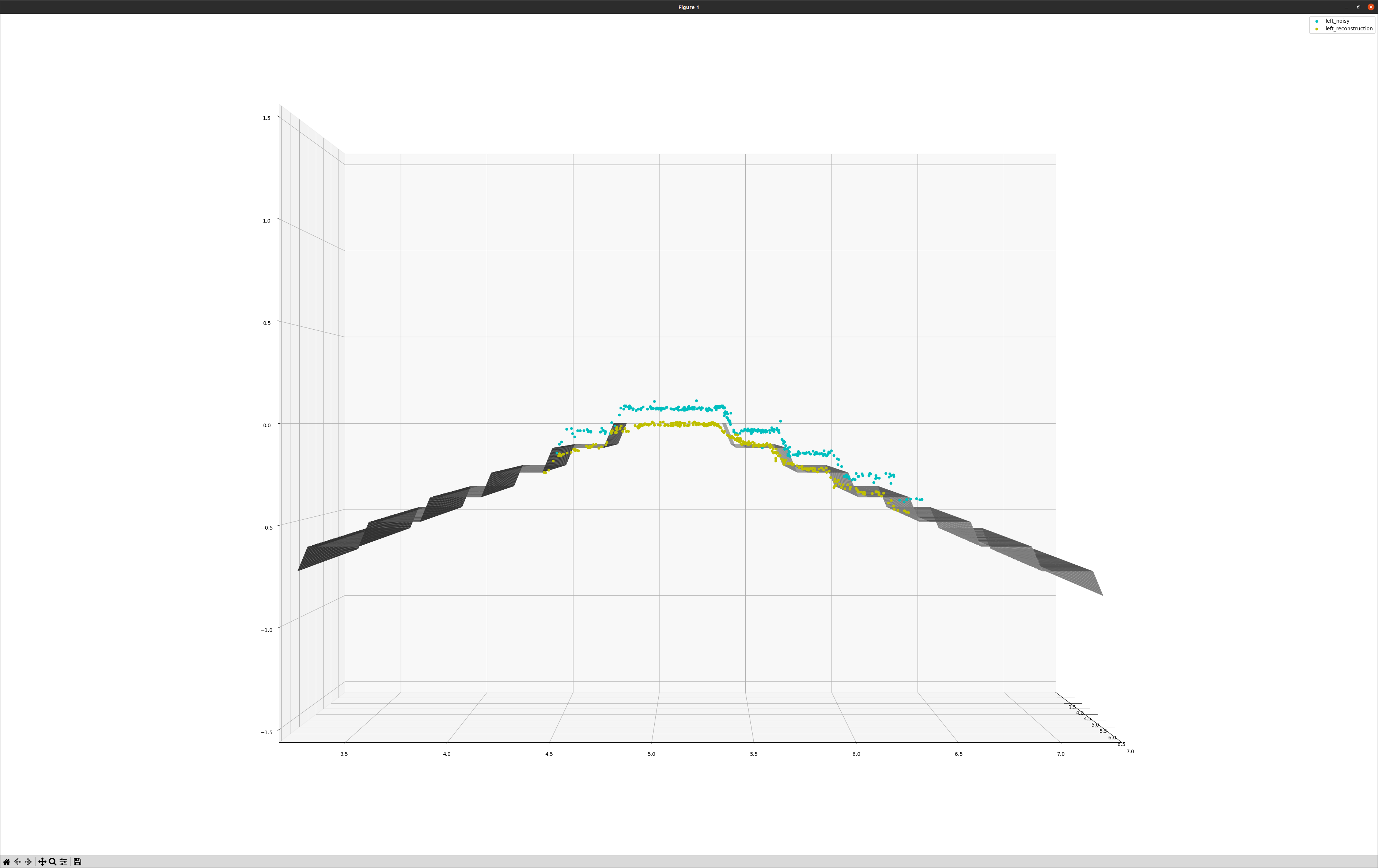}}
		\caption{A side view of the sampled height map and reconstructed height map for the left foot during stair traversal. Cyan is the sampled noisy height map, yellow is the reconstruction produced by the belief decoder. The policy is able to denoise large offsets in the height map.}
		\label{fig:denoising}
\end{figure}

An important aspect of the belief encoder system is the ability to interpret noisy exteroceptive inputs along with proprioception to form an accurate belief of the environment. In order to demonstrate this capability we show a view of the exteroceptive reconstruction produced by the belief decoder in Figure~\ref{fig:denoising}. Our policy is able to denoise large, sometimes alternating offsets in the height map. Additionally, the belief encoder is able to eliminate outliers, while keeping an accurate representation of the terrain.
	
\section{Discussion}
\label{sec:discussion}
In this work we presented a method to learn a bipedal locomotion policy that can utilize exteroceptive observations to successfully traverse irregular terrain, while following a command. We have shown that such a policy observing exteroception greatly outperforms a purely proprioception based locomotion policy when traversing irregular terrains. An exteroceptive policy is able to achieve this outperformance on terrain while at the same time increasing speed, stability and energy efficiency. Critically, we have shown that our policy has learned to achieve such behavior while relying on noisy exteroceptive observations, showcasing the robustness of the control policy.

Limitations of this work include the fact that our experiments have only been conducted in simulation. Although most methods used in this work have been proven to work on real robots in the past, future work should include testing on a real robot to confirm results.
Another limitation we observe is the low speed of iteration in reward and curriculum design caused by the multiple day training time of the policy. Future work could explore the use of faster training methods to more effectively optimize the rewards and curriculum, such as presented in~\cite{rudin2021parallel}.


\section*{Acknowledgment}
This work was done as part of the Master's thesis of the first author. We thank the authors of the \textit{cassie-mujoco-sim} environment~\cite{cassieMujocoSim} for making their code publicly available. Finally, we thank the Center for Information Technology of the University of Groningen for their support and for providing access to the Peregrine high performance computing cluster.

\bibliographystyle{IEEEtran}
\bibliography{biblio}
\end{document}